\documentclass{article}

\PassOptionsToPackage{numbers, compress}{natbib}
 \usepackage[preprint]{neurips_2026}

\usepackage[utf8]{inputenc} 
\usepackage[T1]{fontenc}    
\usepackage{hyperref}       
\usepackage{url}            
\usepackage{booktabs}       
\usepackage{amsfonts}       
\usepackage{nicefrac}       
\usepackage{microtype}      
\usepackage{xcolor}         

\usepackage{amsmath}
\usepackage{multirow}
\usepackage{graphicx}
\usepackage{subcaption}
\usepackage{caption}
\usepackage{float}
\usepackage{tikz}
\usepackage[table]{xcolor}
\usepackage{wrapfig}
\newcommand{\modelName}{SteadyDancer}
\newcommand{\dataName}{X-Dance}

\title{\modelName: Harmonized and Coherent Human \\ Image Animation with First-Frame Preservation}

\author{%
  Jiaming Zhang$^{1,\ddag}$\thanks{Work is done during internship at Tencent PCG. \quad
  $\ddag$ Equal contribution. \quad
  $\dag$ Corresponding author (lmwang@nju.edu.cn).} \quad
  Shengming Cao$^{2,\ddag}$ \quad
  Rui Li$^{2,\ddag}$ \quad
  Xiaotong Zhao$^2$ \quad
  Yutao Cui$^2$ \\
  Xinglin Hou \quad
  Gangshan Wu$^1$ \quad
  Haolan Chen$^2$ \quad
  Yu Xu$^2$ \quad
  Limin Wang$^{1,3,\dag}$ \quad
  Kai Ma$^{2,\dag}$ \\
  $^1$State Key Laboratory for Novel Software Technology, Nanjing University\\
  $^2$Platform and Content Group (PCG), Tencent \quad $^3$Shanghai AI Lab\\
}

\begin{document}

\maketitle

\begin{figure}[H]
\centering
\vspace{-8mm}
\includegraphics[width=0.85\textwidth]{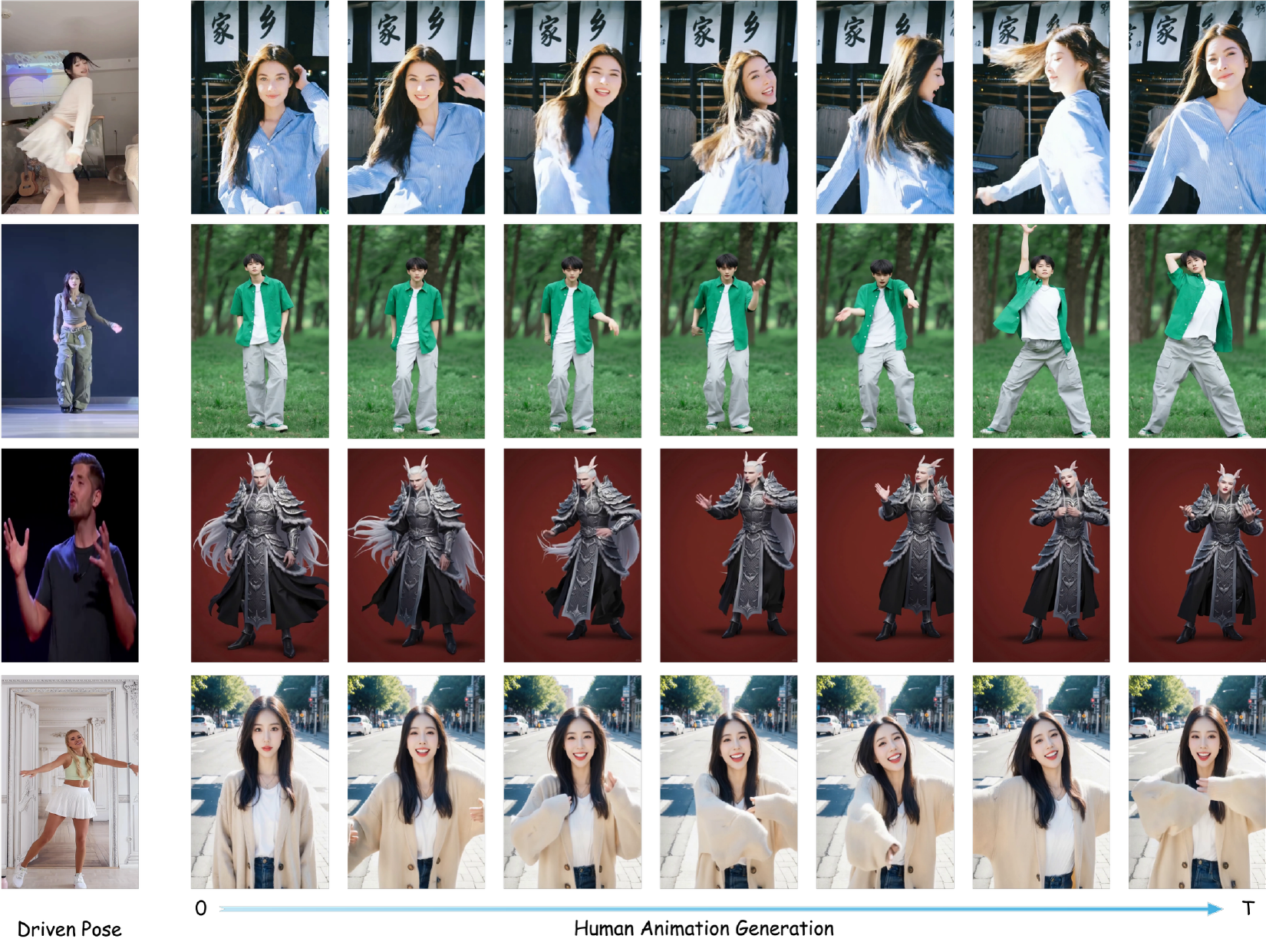}
\vspace{-2mm}
\caption{We introduce \modelName, an Image-to-Video (I2V) paradigm animation framework to achieve harmonized and coherent animation with first-frame preservation.}
\label{first_figure}
\vspace{-6mm}
\end{figure}

\begin{abstract}
Preserving first-frame identity while ensuring precise motion control is a fundamental challenge in human image animation.
The Image-to-Motion Binding process of the dominant Reference-to-Video (R2V) paradigm overlooks critical spatio-temporal misalignments common in real-world applications, leading to failures such as identity drift and visual artifacts.
We introduce SteadyDancer, an Image-to-Video (I2V) paradigm-based framework that achieves harmonized and coherent animation and is the first to ensure first-frame preservation robustly.
Firstly, we propose a Condition-Reconciliation Mechanism to harmonize the two conflicting conditions, enabling precise control without sacrificing fidelity.
Secondly, we design Synergistic Pose Modulation Modules to generate an adaptive and coherent pose representation that is highly compatible with the reference image.
Finally, we employ a Staged Decoupled-Objective Training Pipeline that hierarchically optimizes the model for motion fidelity, visual quality, and temporal coherence.
Experiments demonstrate that SteadyDancer achieves state-of-the-art performance in both appearance fidelity and motion control, while requiring significantly fewer training resources than comparable methods.
The model has been publicly released at \url{https://mcg-nju.github.io/steadydancer-web}.
\end{abstract}
\section{Introduction}
\label{sec:intro}

Human image animation, which aims to generate video from a single static image with controllable motion, has emerged as a prominent research frontier in video generation and holds immense potential for applications like film production, advertising, and video game development.
While significant progress has been made, the breakthroughs of diffusion models~\cite{HunyuanVideo, Wan21} have recently unlocked new capabilities in generating high-fidelity and temporally coherent videos.

\begin{figure}[t]
    \centering
    \begin{subfigure}[b]{0.37\linewidth}
        \centering
        \includegraphics[width=\linewidth]{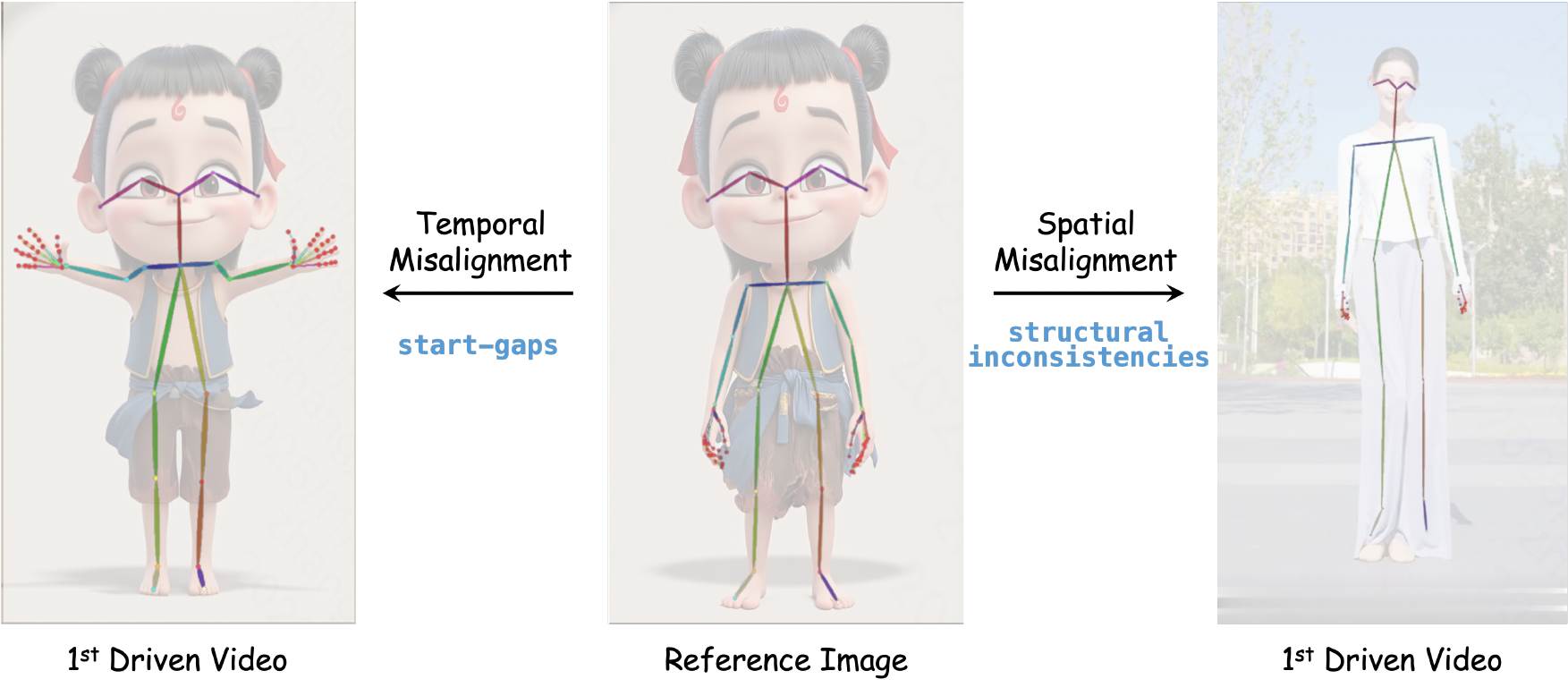}
        \caption{} 
        \label{fig:task_mismatch}
    \end{subfigure}%
    \begin{subfigure}[b]{0.015\linewidth}
        \centering
        \begin{tikzpicture}
            \draw[dashed, line width=1pt] (0,0) -- (0, 3.4); 
        \end{tikzpicture}
    \end{subfigure}%
    \begin{subfigure}[b]{0.415\linewidth}
        \centering
        \includegraphics[width=\linewidth]{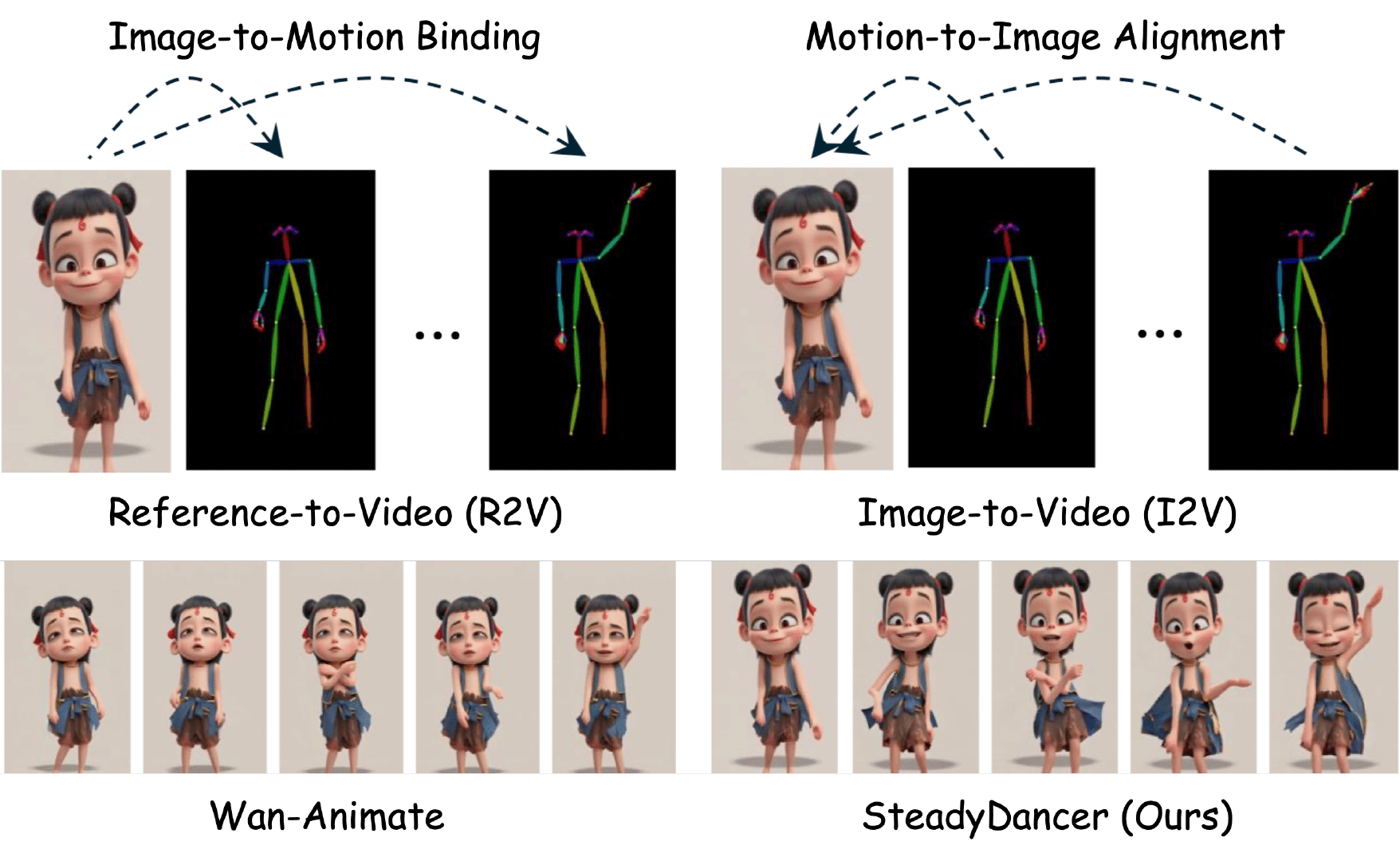}
        \caption{}
        \label{fig:task_R2V_I2V}
    \end{subfigure}
    \vspace{-2mm}
    \caption{(a) The spatio-temporal misalignment in practical scenarios. (b) The illustration of Reference-to-Video (R2V) and Image-to-Video (I2V) paradigms for human image animation. While R2V only cares about how to \textit{bind} the reference image to driven motion, I2V additionally needs to carefully deal with the \textit{misalignment} from the driven motion to the reference image.}
    \label{fig:two-task}
    \vspace{-6mm}
\end{figure}
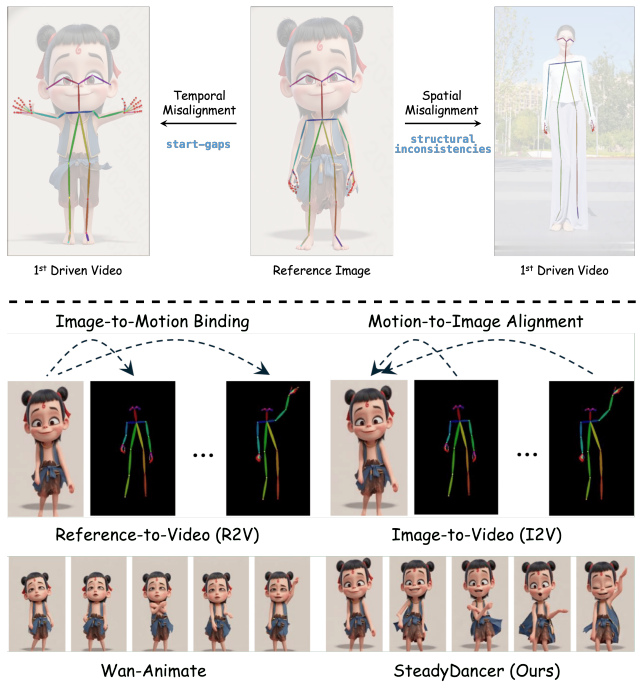

Most existing methods adhere to the Reference-to-Video (R2V) paradigm~\cite{vace}, as illustrated in the left part of Fig.~\ref{fig:two-task} (b).
Specifically, the R2V paradigm defines the animation task as binding the reference image onto the driven pose, which inherently relaxes the alignment constraints between the inputs, thereby reducing the objective to only achieving a natural appearance transfer.
However, inherent discrepancies exist between the image and pose inputs in practical scenarios as shown in Fig.~\ref{fig:two-task} (a), manifesting as \textit{spatial misalignments} limb structure and proportion) and \textit{temporal misalignments} (movement type and amplitude).
In such scenarios, the relaxation of alignment constraints within the R2V paradigm leads to unacceptable results, including severe artifacts, appearance distortions, and temporal incoherence.
Moreover, in the common \textit{start-gap} scenario due to the temporal misalignments, the R2V paradigm instantly binds the reference image to the first pose, completely omitting any smooth transition from the reference state.
In real-world applications, including VFX post-production, keyframe-based animation, and digital human activation, the generated video must start exactly from the reference frame. However, this abrupt jump, combined with the appearance distortion, makes it unsuitable for these applications that require high visual and temporal fidelity.

In contrast, the Image-to-Video (I2V) paradigm, which inherently guarantees \textit{first-frame preservation} by generating consistent and coherent videos starting from the initial frame, maximizes fidelity and emerges as a preferable solution.
However, I2V-based animation research remains scarce and technically challenging, due to its stringent requirement for first-frame coherence, which demands that all input conditions and generated results adhere to the initial frame.
Specifically, it requires the pose to be modulated into a suitable control signal for the reference image, necessitating much tighter alignment than R2V.
Failure to achieve such strict alignment will severely impact performance, especially when there are insufficient resources to train high-capacity models for learning control.

In this paper, we propose \textbf{\textit{\modelName}}, an animation model built on the I2V paradigm with first-frame preservation.
Firstly, to resolve the trade-off between appearance preservation and motion control within the I2V paradigm, we propose \textit{Condition-Reconciliation Mechanism} that achieves precise motion-driven control without sacrificing first-frame fidelity by optimizing at three levels, including condition fusion, injection, and augmentation.
Secondly, to address the spatio-temporal misalignment between the reference image and the driving pose, we introduce \textit{Synergistic Pose Modulation Modules} to extract pose representations that are better adapted to the reference image, comprising the Spatial Structure Adaptive Refiner, the Temporal Motion Coherence Module, and the Frame-wise Attention Alignment Unit.
Finally, to achieve efficient and stable model training, we propose a \textit{Staged Decoupled-Objective Training Pipeline}, including an Action Supervision Stage to establish precise motion control, a Condition-Decoupled Distillation Stage to enhance the generated video quality, and a Motion Discontinuity Mitigation Stage that aims to generate smooth and continuous results.
Ultimately, building on these strategies, \modelName~successfully activates the ability of first-frame preservation in the human animation task, enabling it to robustly handle misalignments and generate videos with high fidelity and accurate motion.
Moreover, it achieves superior performance over existing methods while requiring \textit{substantially fewer training resources}.
Meanwhile, we propose \textit{the first non-homogeneous benchmark, \textbf{\textit{\dataName}}}, to test model performance when the reference and action sources are inconsistent.

In summary, the main contribution of this work lies in:
(i) We propose \modelName, a novel high-fidelity animation framework that achieves first-frame preservation firstly and significant training resource efficiency;
(ii) To address the conflict and mismatch between the reference image and the driven pose, and to improve training efficiency, we propose a Condition-Reconciliation Mechanism, Synergistic Pose Modulation Modules, and a Staged Decoupled-Objective Training Pipeline;
(iii) Extensive quantitative and qualitative results on multiple benchmarks validate the superiority and effectiveness of our proposed method.

\section{Related Work}\label{sec:rel-work}

\noindent \textbf{Diffusion for video generation}
Diffusion-based models have become state-of-the-art for generative tasks, achieving remarkable success in both image~\cite{DBLP:conf/cvpr/RombachBLEO22,ControlNet} and video generation~\cite{MAGVIT,DBLP:journals/corr/abs-2310-19512}.
With the introduction of OpenAI's Sora~\cite{sora}, which leverages the Diffusion Transformer (DiT) architecture~\cite{DiT}, DiT-based approaches have since supplanted UNets~\cite{unet}, as their pure Transformer structure facilitates massive parameter scaling and has become the mainstream technical route. Concurrently, to efficiently handle video data, many recent DiT models~\cite{Wan21,HunyuanVideo,CogVideoX} adopt 3D VAEs over standard 2D VAEs~\cite{DBLP:conf/cvpr/RombachBLEO22,VAE} to compress data across both spatial and temporal dimensions.
The proliferation of powerful, open-source foundational models has further accelerated this field. Consequently, human image animation, as one of downstream tasks, directly benefits from the enhanced power and fidelity of these new models.

\noindent \textbf{Human Image Animation.}
Earlier works in image animation primarily relied on warping-based feature representations and GAN-based architectures~\cite{siarohin2019first,siarohin2021motion,zhao2022thin}.
Recently, this field has pivoted to diffusion models, yielding significant performance improvements.
Early diffusion-based methods, such as DisCo~\cite{wang2024disco}, leveraged ControlNet~\cite{ControlNet} for pose guidance and integrated motion modules to enhance cross-frame consistency.
A key breakthrough came with Animate Anyone~\cite{AnimateAnyone} and subsequent studies~\cite{MagicAnimate,zhu2024champ,tu2025stableanimator}, which utilize a UNet-based ReferenceNet to inject appearance features, achieving excellent identity preservation.
To further enhance controllability, other works~\cite{zhu2024champ,hu2025animateanyone2,zhou2024realisdance,HumanVid,MimicMotion} incorporated camera parameters and rich 3D geometric guidance, such as depth, SMPL, hand.
Mirroring the trend in general video generation, DiT-based architectures~\cite{DiT} have recently been adapted for human animation~\cite{zhang2025flexiact,HyperMotion,UniAnimate-DiT,RealisDance-DiT,luo2025dreamactorm1,Wan-Animate,vace}, leading to substantial enhancements in realism and temporal continuity.
However, most approaches follow the Reference-to-Video (R2V) paradigm. This paradigm focuses on binding Image-to-Pose naturally, which inherently ignores critical input misalignments, frequently leading to unsatisfactory results.
\section{Method}

\begin{figure}[!t]
\centering
\includegraphics[width=1.0\textwidth]{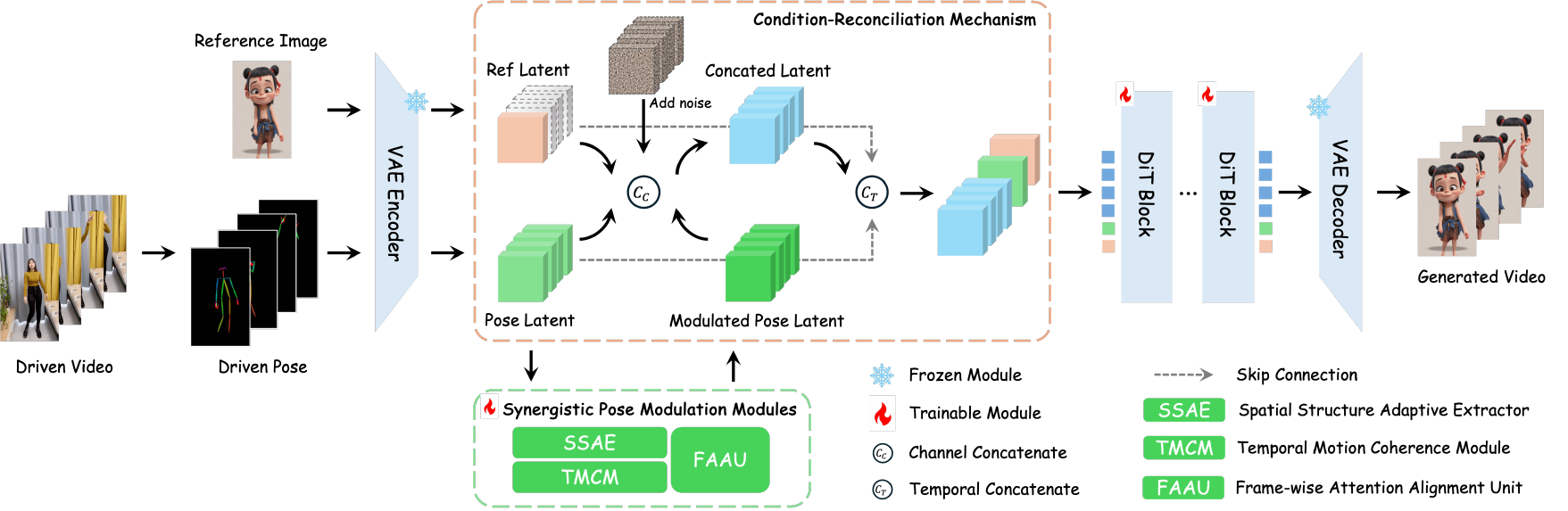}
\vspace{-5mm}
\caption{An overview of \modelName, a Image-to-Video (I2V) paradigm-based Human Image Animation framework. First, it employs a \textit{Condition-Reconciliation Mechanism} to reconcile appearance and motion conditions, achieving precise control without sacrificing \textit{first-frame preservation}. Second, it utilizes Synergistic \textit{Pose Modulation Modules} to resolve critical \textit{spatio-temporal misalignments}.}
\label{fig:framework}
\vspace{-5mm}
\end{figure}

Given a reference image $I_{c}$ along with a pose sequence $P_{m}=\{p_{0}, \dots, p_{T}\}$, the animation task aims to generate a video that preserves the appearance of the reference while maintaining adherence to the pose sequence.
As shown in Fig.~\ref{fig:framework}, we first review the Image-to-Video (I2V) generation model (Sec.~\ref{sec:method-Preliminaries}) and introduce a Na\"ive I2V Baseline to highlight its limitations (Sec.~\ref{sec:method-naiveBaseline}).
We then present the core technical components, including the Condition-Reconciliation Mechanism (Sec.~\ref{sec:method-cond}) and the Synergistic Pose Modulation Modules (Sec.~\ref{sec:method-posemodules}).
Finally, we describe the Staged Decoupled-Objective Training Pipeline (Sec.~\ref{sec:method-traing}).

\subsection{Preliminaries}\label{sec:method-Preliminaries}

A foundational Image-to-Video (I2V) model~\cite{Wan21,CogVideoX,SVD} conditions synthesis on a static image $I_c$, used as the first frame, concatenated with zero-filled frames, and encoded by a VAE encoder $\mathcal{E}$~\cite{VAE} into $z_c$.
At denoising timestep $t$, the Diffusion Transformer (DiT)~\cite{DiT} receives the channel-concatenated noisy latent $\hat{z}_t$, binary mask $m$, and condition latent $z_c$:
\vspace{-3mm}
\begin{equation}
    z_t = \text{ChannelConcat}(\hat{z}_t, m, z_c).
\end{equation}
The DiT predicts the denoised latent from $z_t$, while global context $c_{clip}$ and text conditions $c_{txt}$ are injected through decoupled cross-attention for spatial-semantic alignment.

\subsection{Na\"ive I2V Baseline}\label{sec:method-naiveBaseline}
To introduce pose control $P_{m}$, we build a Na\"ive Baseline that treats pose and image conditions equivalently.
The pose sequence reuses the image VAE encoder to obtain $z_p$ in the same feature space as $z_c$, then fuses them by element-wise addition for denoising:
\vspace{-1mm}
\begin{equation}
    z_t = \text{ChannelConcat}(\hat{z}_t, m, z_c + z_p).
\end{equation}
, which achieves both appearance preservation and motion control by simple element-wise addition.

\subsection{Condition-Reconciliation Mechanism}\label{sec:method-cond}
The Na\"ive Baseline uses simple additive fusion to preserve appearance and motion, but this conflates static image details with driving-pose dynamics, a conflict especially acute in I2V.
The fused signal often favors control and weakens appearance retention.
We therefore propose the Condition-Reconciliation Mechanism, a three-aspect design for precise control with first-frame preservation.

\noindent\textbf{Condition Fusion.} 
We identify that the element-wise addition is a critical bottleneck of Naïve Baseline, which conflates the static appearance ($z_c$) and dynamic pose ($z_p$) signals, leading to information loss and mutual interference.
To resolve this, we replace it with channel-wise concatenation as:
\vspace{-2mm}
\begin{equation}
    z_t = \text{ChannelConcat}(\hat{z}_t, m, z_c, z_p),
\end{equation}which keeps conditions distinct, improving appearance preservation and motion control.

\noindent\textbf{Condition Injection.}
Parameter-intensive injection, e.g., adapters or decoupled cross-attention, increases trainable parameters and attention cost, risking interference with the pre-trained generator.
We instead inject the pose latent $z_p$ with the image condition and apply LoRA fine-tuning, enhancing motion control while preserving generation capacity and first-frame fidelity.

\noindent\textbf{Condition Augmentation.}
To further reinforce the preservation of the first frame,  we introduce two augmentation strategies.
First, we augment the DiT input at the temporal level. We take the channel-concatenated latent $z_{\text{cond}}$ from the Condition Fusion and temporally concatenate it with the image latent $z_{c_0}$ and the first-frame pose latent $z_{p_0}$ as:
\vspace{-2mm}
\begin{equation}
\begin{aligned}
    z_{\text{cond}} &= \text{ChannelConcat}(\hat{z}_t, m, z_c, z_p), \\
    z_t &= \text{TemporalConcat}(z_{\text{cond}}, z_{c_0}, z_{p_0}).
\end{aligned}
\label{eq:Augmentation}
\end{equation}
This provides the model with an explicit, clean reference to the starting appearance and pose.
Second, we enhance the global context $c_{clip}$ by concatenating it with the CLIP feature of the first pose frame. This provides the model with a richer, pose-aware semantic embedding. These two augmentations work synergistically to improve identity preservation and visual consistency.

\subsection{Synergistic Pose Modulation Modules}\label{sec:method-posemodules}
While our Condition-Reconciliation Mechanism improves the fidelity-control balance (Fig.~\ref{fig:two-task}), spatio-temporal misalignment remains.
\textit{Spatial misalignment} arises from source-pose disparities in skeleton or identity, causing structural changes, identity drift, and detail loss.
\textit{Temporal misalignment} stems from noisy poses and abrupt source-to-pose transitions, producing jitter and realism degradation; we therefore design modulation modules for precise alignment.

\noindent\textbf{Spatial Structure Adaptive Extractor.}
To address spatial misalignment, we propose the Spatial Structure Adaptive Refiner $\mathcal{P}_{SSAE}$ with \textit{dynamic convolution}, which generates pose-dependent parameters from $z_p$ to modulate latent representations.
Using global context to predict scaling factors and a transformation matrix, its dual-path restructuring produces image-compatible pose features that reduce fusion interference and preserve fine details.

\noindent\textbf{Temporal Motion Coherence Module.}
For temporal misalignment, we introduce the Temporal Motion Coherence Module $\mathcal{P}_{TMCM}$ to model continuous dynamics from the discrete pose $z_{p}$.
Stacked factorized blocks use \textit{depthwise spatial convolutions} for intra-frame structures and \textit{pointwise temporal convolutions} for smoothed inter-frame dynamics.
This suppresses erratic-pose artifacts and yields coherent motion guidance.

\noindent\textbf{Frame-wise Attention Alignment Unit.}
To enforce fine-grained pose-appearance correspondence, we introduce the lightweight Frame-wise Attention Alignment Unit $\mathcal{P}_{FAAU}$ via cross-attention, where denoising latent $\hat{z}_t$ (\textit{Query}) attends to pose latent (\textit{Key / Value}).
It yields appearance-calibrated pose features for subsequent fusion.

\noindent\textbf{Hierarchical Aggregation.}
Ultimately, we combine the aforementioned three modules using a hierarchical aggregation strategy.
First, the base pose feature is processed in parallel by the spatial ($\mathcal{P}_{SSAE}$) and temporal ($\mathcal{P}_{TMCM}$) modules. Their outputs are then integrated with the base feature via a residual connection to construct a high-quality, spatio-temporally coherent representation. This intermediate representation is immediately calibrated by the Alignment Unit ($\mathcal{P}_{FAAU}$).
This synergistically refined pose feature achieves precise appearance alignment, which then serves as an additional, high-quality control condition for Eq.~\ref{eq:Augmentation}.
The aggregation process can be formalized as:
\vspace{-2mm}
\begin{equation}
\begin{aligned}
    z_{p^*} &= z_p + \mathcal{P}_{SSAE}(z_p) + \mathcal{P}_{TMCM}(z_p), \\
    z_{p^{\dag}} &= \mathcal{P}_{FAAU}(q=\hat{z}_t, kv=z_{p^*}), \\
    z_{\text{cond}} &= \text{ChannelConcat}(\hat{z}_t, m, z_c, z_{p^*}, z_{p^{\dag}}). \\
\end{aligned}
\vspace{-3mm}
\end{equation}

\subsection{Staged Decoupled-Objective Training Pipeline}\label{sec:method-traing}

Our training pipeline is divided into three distinct stages to achieve efficient and precise training.

\noindent\textbf{Action Supervision.}
This stage instills motion-control capability.
For each training video, the first frame is reference, while the full video provides the motion condition and supervision target.
We use LoRA-based~\cite{LoRA} fine-tuning to preserve the model's generative priors.

\noindent\textbf{Condition-Decoupled Distillation.}\label{sec:method-training-cdd}
To recover quality lost during motion-control learning, the second stage aims to improve detail while retaining first-stage pose controllability.
We use the original pre-trained I2V model as the \textit{teacher model $\theta$}, a stationary manifold for \textit{the unconditional data distribution}, and the first-stage model as the \textit{student model $\phi$}, a \textit{conditional flow estimator}.
We decompose velocity prediction into unconditional and conditional components:
\vspace{-2mm}
\begin{equation}
    \mathbf{v}^{\phi}(\mathbf{x}_t, t, \mathbf{c}) = \underbrace{\mathbf{v}_{\text{u}}^{\phi}(\mathbf{x}_t, t)}_{\text{unconditional component}} + \underbrace{\mathbf{v}_{\text{c}}^{\phi}(\mathbf{x}_t, t, \mathbf{c})}_{\text{conditional component}},
\end{equation}
where $\mathcal{L}_{\text{distill}} = \lVert \mathbf{v}_{\text{u}}^{\phi}(\mathbf{x}_t, t) - \mathbf{v}_{\text{u}}^{\theta}(\mathbf{x}_t, t) \rVert^{2}$ aligns the unconditional component with the frozen teacher, and $\mathcal{L}_{\text{fidelity}} = \lVert \mathbf{v}^{\phi}(\mathbf{x}_t, t, \mathbf{c}) - \mathbf{v}^{*} \rVert^{2}$ regresses the ground-truth velocity field $\mathbf{v}^{*}$ like the first stage.
Consequently, the teacher's unconditional manifold is injected into the student without biasing pose-specific conditions, eliminating the distribution shift when a conditional network mimics an unconditional target, thereby improving video quality.

\begin{figure}[t]
    \centering
    \begin{minipage}[t]{0.48\textwidth}
        \centering
        \includegraphics[width=\linewidth]{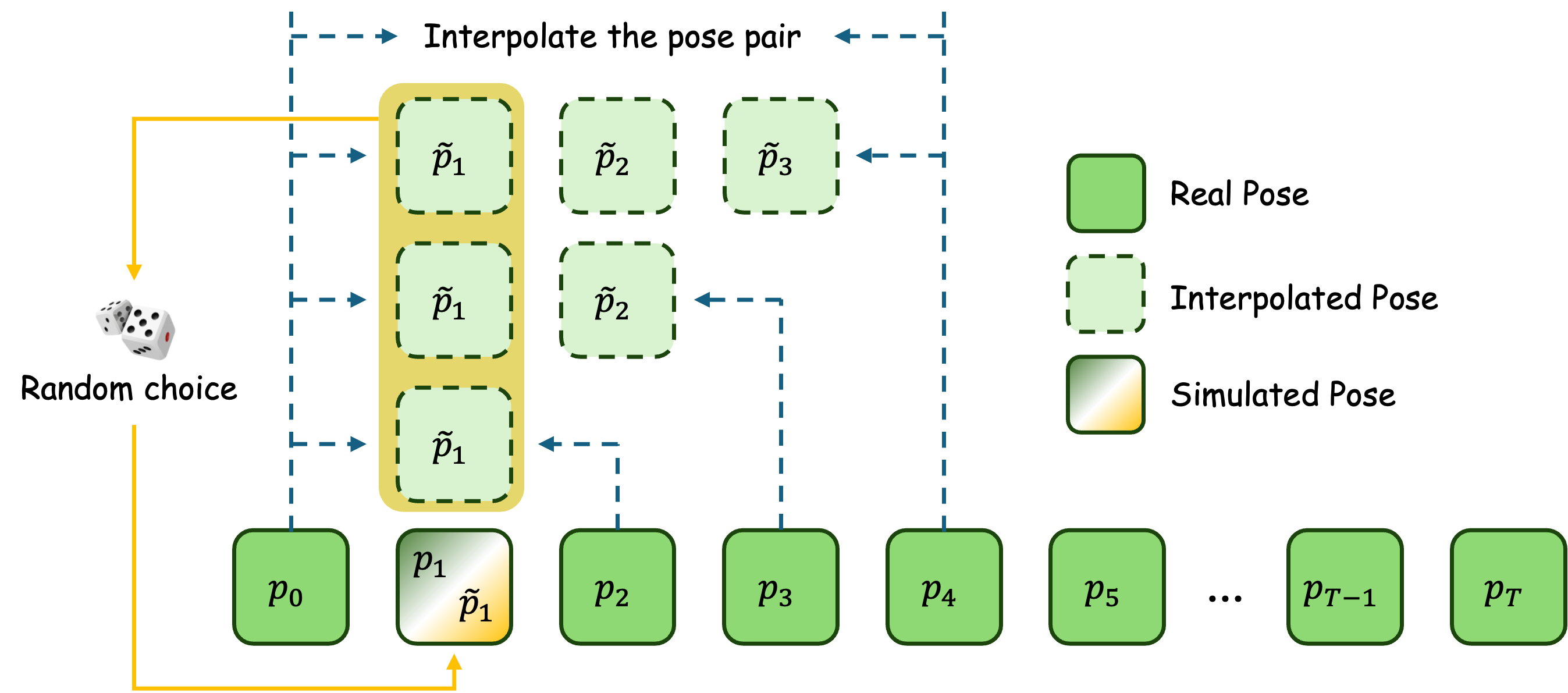}
        \caption{Pose Simulation in Motion Discontinuity Mitigation of Staged Decoupled-Objective Training Pipeline, which randomly replaces the second pose with an interpolated pose.}
        \label{fig:pose_imitation_process}
    \end{minipage}
    \hfill 
    \begin{minipage}[t]{0.48\textwidth}
        \centering
        \includegraphics[width=\linewidth]{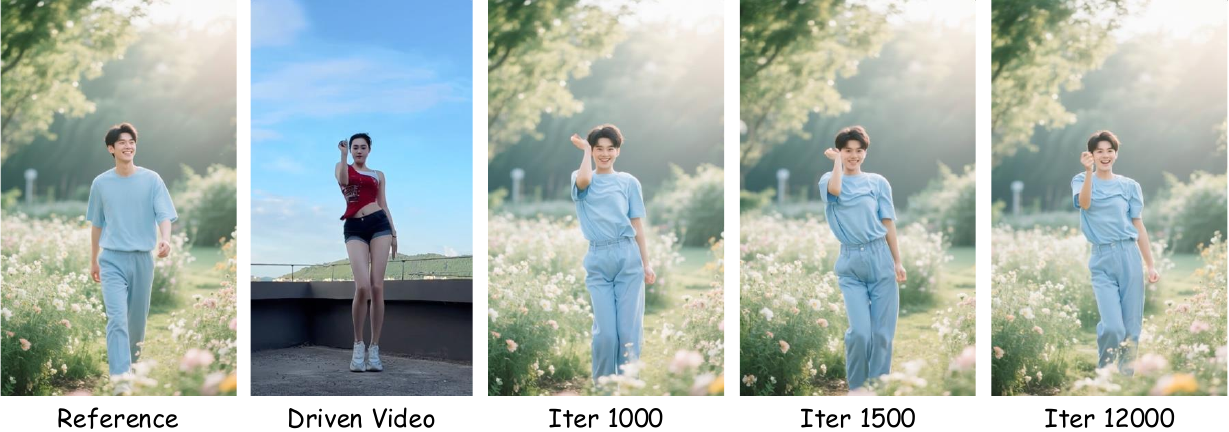}
        \caption{Model performance across various training steps. The results indicate that it rapidly acquires motion-control in the early steps, while the later steps focus more on detail.}
        \label{fig:ablation_ckpt}
    \end{minipage}
    \vspace{-5mm}
\end{figure}

\noindent\textbf{Motion Discontinuity Mitigation.}\label{sec:method-training-mdas}
At test time, reference--pose gaps can cause abrupt start-frame jumps.
Training lacks this start-gap because the reference image and first pose are identical ($v_{0}$), while random jumps destabilize learning.
We therefore propose \textit{Pose Simulation} (Fig.~\ref{fig:pose_imitation_process}): for a smooth sequence $\{p_{0}, p_{1}, \dots, p_{T}\}$, we sample $(p_{0}, p_{T^{*}})$ with $T^{*}\in\{2, 3, 4\}$, interpolate $\{\tilde{p}_{1}, \dots, \tilde{p}_{T^{*}-1}\}$, and replace ${p}_{1}$ with $\tilde{p}_1$ to form $\{p_{0}, \tilde{p}_1, \dots, p_{T}\}$.
This strategy needs only a few hundred fine-tuning steps and no architectural changes, resolving over 80\% of discontinuities without post-processing.

\section{Experiments}

\subsection{Experimental Setups}

\begin{table}[!t]
    \centering
    \caption{Comparison of Extra Inputs, Pre-Trained Model, and Training Requirements.}
    \label{tab:train_resource}
    \resizebox{\linewidth}{!}{
        \begin{tabular}{cccccc}
\toprule
Method & Extra Inputs & Pre-Trained Model          & Training Step                 & Training Data                           \\
\midrule
\multicolumn{1}{l}{\textbf{\textit{UNet-based}}} &                    &                            &                               &                                         \\
Disco {\tiny\textcolor{gray}{[CVPR24]}}~\cite{wang2024disco} & Mask & Stable Diffusion           & 70k                           & TikTok (350)                            \\
Animate Anyone {\tiny\textcolor{gray}{[CVPR24]}}~\cite{AnimateAnyone} & $\times$   & Stable Diffusion           & {[}30k, 10k{]}                & 5k character video clips                \\
MagicAnimate {\tiny\textcolor{gray}{[CVPR24]}}~\cite{MagicAnimate} & DensePose          & Stable Diffusion           & -                             & TikTok (350), TED-talks (1,203)         \\
Champ {\tiny\textcolor{gray}{[ECCV24]}}~\cite{zhu2024champ} & Depth, SMPL        &  Stable Diffusion                          & {[}60k, 20k{]}                & 5k human videos \\
HumanVid {\tiny\textcolor{gray}{[NeurIPS24]}}~\cite{HumanVid} & Camera             & Stable Diffusion           & {[}30k, 10k{]}                & 20k real, 50k synthetic                                     \\
RealisDance {\tiny\textcolor{gray}{[arxiv24]}}~\cite{zhou2024realisdance} & SMPL, HaMeR        & Real Vision                & {[}200k, 100k{]}              & about 16k videos                        \\
StableAnimator {\tiny\textcolor{gray}{[CVPR25]}}~\cite{tu2025stableanimator} & Face               & Stable Video Diffusion     & 20 ep ($\sim$15k)         & 3K videos (60-90 seconds long)          \\
X-Dyna {\tiny\textcolor{gray}{[CVPR25]}}~\cite{X-Dyna} & Face               & Stable Diffusion           & {[}5ep, 2ep{]} & 107,546, 30-second videos               \\
MIMO {\tiny\textcolor{gray}{[CVPR25]}}~\cite{MIMO} & Depth, Mask        & Stable Diffusion           & 50K                           & 5k real, 2k synthetic videos                   \\
Animate-X {\tiny\textcolor{gray}{[ICLR25]}}~\cite{Animate-X} & $\times$   & Stable Diffusion           & -                             & 9k human videos                         \\
MimicMotion {\tiny\textcolor{gray}{[ICML25]}}~\cite{MimicMotion} & $\times$   & Stable Video Diffusion & 20 ep ($\sim$11k)         & 4,436 human dancing videos              \\
\midrule
\multicolumn{1}{l}{\textbf{\textit{DiT-based}}}  &                    &                            &                               &                                         \\
Dreamactor-M1 {\tiny\textcolor{gray}{[ICCV25]}}~\cite{luo2025dreamactorm1} & Face               & Seaweed APTs               & {[}20k, 20k, 30k{]}           & 500-hour videos                               \\
VACE {\tiny\textcolor{gray}{[ICCV25]}}~\cite{vace} & Mask               & Wan-2.1 T2V 14B            & 200k                          & 1M                                      \\
FlexiAct {\tiny\textcolor{gray}{[SIGGRAPH25]}}~\cite{zhang2025flexiact} & $\times$   & CogVideoX-I2V              & {[}40K, 1.5K{]}               & -                                       \\
X-UniMotion {\tiny\textcolor{gray}{[SIGGRAPH ASIA25]}}~\cite{X-UniMotion} & Face, Hands   & Seaweed-7b                 & 40k                           & 200-hour videos                               \\
UniAnimate-DiT {\tiny\textcolor{gray}{[arxiv25]}}~\cite{UniAnimate-DiT} & $\times$   & Wan-2.1 I2V 14B            & -                           & $\sim$10K human dance videos            \\
RealisDance-DiT {\tiny\textcolor{gray}{[arxiv25]}}~\cite{RealisDance-DiT} & SMPL               & Wan-2.1 I2V 14B            & -                             & 1M high-quality videos                  \\
HyperMotion {\tiny\textcolor{gray}{[arxiv25]}}~\cite{HyperMotion} & $\times$   & Wan-2.1 I2V 14B            & 20k                           & 19,597 video clips  \\
Wan-Animate {\tiny\textcolor{gray}{[arxiv25]}}~\cite{Wan-Animate} & Face               & Wan-2.1 I2V 14B            & -                             & -                                       \\
\rowcolor[rgb]{0.929,0.902,0.973}
\modelName~(Ours) & $\times$   & Wan-2.1 I2V 14B            & {[}12k, 2k, 0.5k{]}                           & 7338 videos (10.2 hours)               \\
\bottomrule
\end{tabular}
    }
    \vspace{-6mm}
\end{table}

\begin{figure}[t]
    \centering
    \includegraphics[width=\textwidth]{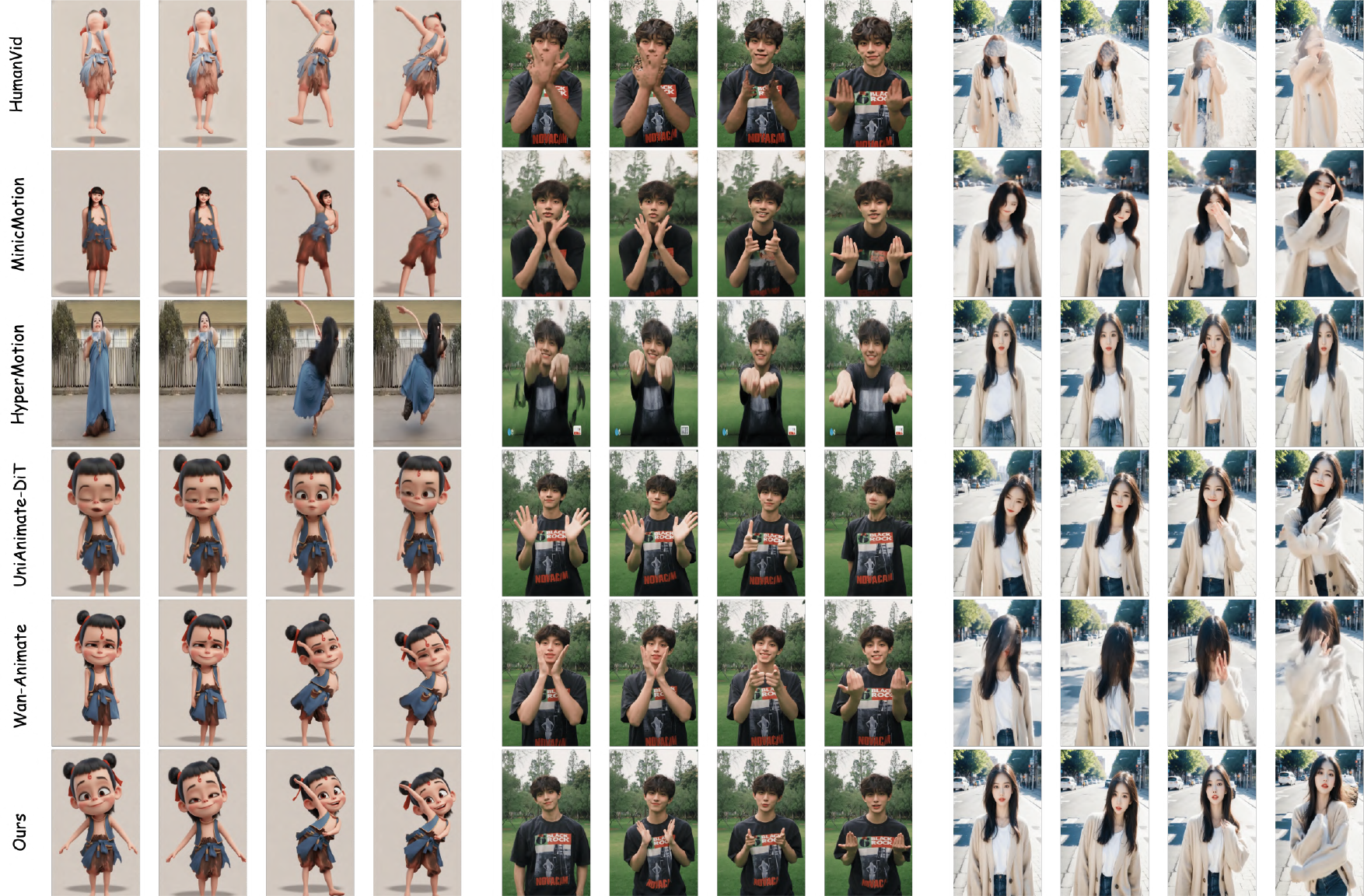}
    \caption{Qualitative comparisons between SteadyDancer and other methods on the X-Dance. Each example displays the evolution (starting from the first frame), highlighting our model's superior \textbf{high-fidelity, coherence, and first-frame preservation}.}
    \label{fig:compare_I2V}
    \vspace{-3mm}
\end{figure}

\begin{table}[t]
  \caption{Quantitative Results on the TikTok, RealisDance-Val, and our X-Dance. .}
  \label{tab:tiktok_realisedance}
  \centering
  \begin{subtable}{0.2\textwidth}
    \centering
    \small
    \setlength\tabcolsep{3pt}
    \renewcommand\arraystretch{1}
    \resizebox{\textwidth}{!}{\begin{tabular}{lccccc}
\toprule
Method &  SSIM$\uparrow$ & PSNR$\uparrow$ & LPIPS$\downarrow$ & FID$\downarrow$ & FVD$\downarrow$ \\
\midrule
Moore-AnimateAnyone~\cite{AnimateAnyone}    &  0.752   &  16.79   &  0.288   &  52.26   &  935.6  \\
MagicAnimate~\cite{MagicAnimate}    &  0.748   & 17.89   &  0.270   &  56.84   &  876.0  \\
Champ~\cite{zhu2024champ}    &  0.778   &  18.43   &  0.267   &  50.76   &  736.1  \\
Animate-X~\cite{Animate-X}    &  0.740   &  16.71   &  0.280   &  32.77   &  508.6  \\
HumanVid~\cite{HumanVid}    &  0.778   &  18.76   &  0.247   &  41.35   &  691.8  \\ 
Realisdance-DiT~\cite{RealisDance-DiT}    &  0.717   &  17.55   &  0.260   &  30.39   &  458.8  \\
\rowcolor[rgb]{0.929,0.902,0.973}
\modelName~(Ours)    &  0.749   &  17.67   &  0.263   &  30.65   &  451.3  \\

\bottomrule
\end{tabular}}
    \vfill 
    \caption{TikTok dataset.}
    \label{tab:tiktok}
  \end{subtable}
  \hfill
  \begin{subtable}{0.38\textwidth}
    \centering
    \small
    \setlength\tabcolsep{3pt}
    \renewcommand\arraystretch{1}
    \resizebox{\textwidth}{!}{\begin{tabular}{lccccccccc}
\toprule
Model & \multicolumn{1}{c}{\begin{tabular}[c]{@{}c@{}}I2V\\ Subject$\uparrow$\end{tabular}} & \multicolumn{1}{c}{\begin{tabular}[c]{@{}c@{}}I2V\\ Background$\uparrow$\end{tabular}} & \multicolumn{1}{c}{\begin{tabular}[c]{@{}c@{}}Subject\\ Consistency$\uparrow$\end{tabular}} & \multicolumn{1}{c}{\begin{tabular}[c]{@{}c@{}}Background\\ Consistency$\uparrow$\end{tabular}} & \multicolumn{1}{c}{\begin{tabular}[c]{@{}c@{}}Temporal\\ Flickering$\uparrow$\end{tabular}} & \multicolumn{1}{c}{\begin{tabular}[c]{@{}c@{}}Motion\\ Smoothness$\uparrow$\end{tabular}} & \multicolumn{1}{c}{\begin{tabular}[c]{@{}c@{}}Aesthetic\\ Quality$\uparrow$\end{tabular}} & \multicolumn{1}{c}{\begin{tabular}[c]{@{}c@{}}Imaging\\ Quality$\uparrow$\end{tabular}} & \multicolumn{1}{c}{FVD$\downarrow$} \\
\midrule
Moore-AnimateAnyone~\cite{AnimateAnyone} & 94.00             & 94.69  & 94.65       & 94.90          & 97.16       & 98.07     & 51.56     & 66.34   & 748.38    \\
HumanVid~\cite{HumanVid}            & 94.72             & 95.27  & 93.69       & 94.94          & 97.87       & 98.52     & 55.58     & 67.45   & 624.33    \\
MimicMotion~\cite{MimicMotion}         & 90.78             & 92.52  & 92.21       & 93.60          & 97.46       & 98.61     & 52.09     & 59.67   & 785.73    \\
Animate-X~\cite{Animate-X}           & 95.68             & 96.22  & 93.39       & 95.11          & 97.79       & 98.68     & 51.72     & 60.91   & 679.66    \\
Hyper-Motion~\cite{HyperMotion}        & 94.76             & 95.71  & 93.58       & 94.97          & 98.19       & 99.01     & 52.97     & 65.52   & 568.14    \\
UniAnimate-DiT~\cite{UniAnimate-DiT}      & 93.15             & 93.95  & 94.56       & 95.44          & 98.78       & 99.24     & 52.18     & 65.52   & 599.03    \\
VACE~\cite{vace}  & 87.39             & 88.58  & 93.56       & 95.03          & 96.74       & 98.25     & 57.81     & 70.61   & 911.72    \\
Wan-Animate~\cite{Wan-Animate}         & 93.81             & 94.61  & 93.06       & 94.52          & 98.42       & 98.96     & 54.47     & 66.87   & 386.87    \\
\rowcolor[rgb]{0.929,0.902,0.973}
\modelName~(Ours) & 94.68             & 95.38  & 93.48       & 95.18          & 97.99       & 99.02     & 56.80     & 68.45   & 326.49   \\
\bottomrule
\end{tabular}}
    \vfill 
    \caption{RealisDance-Val dataset.}
    \label{tab:realisedance}
  \end{subtable}
  \hfill
  \begin{subtable}{0.38\textwidth}
    \centering
    \small
    \setlength\tabcolsep{3pt}
    \renewcommand\arraystretch{1}
    \resizebox{\textwidth}{!}{\begin{tabular}{lcccccccc}
\toprule
Model & \multicolumn{1}{c}{\begin{tabular}[c]{@{}c@{}}I2V\\ Subject$\uparrow$\end{tabular}} & \multicolumn{1}{c}{\begin{tabular}[c]{@{}c@{}}I2V\\ Background$\uparrow$\end{tabular}} & \multicolumn{1}{c}{\begin{tabular}[c]{@{}c@{}}Subject\\ Consistency$\uparrow$\end{tabular}} & \multicolumn{1}{c}{\begin{tabular}[c]{@{}c@{}}Background\\ Consistency$\uparrow$\end{tabular}} & \multicolumn{1}{c}{\begin{tabular}[c]{@{}c@{}}Temporal\\ Flickering$\uparrow$\end{tabular}} & \multicolumn{1}{c}{\begin{tabular}[c]{@{}c@{}}Motion\\ Smoothness$\uparrow$\end{tabular}} & \multicolumn{1}{c}{\begin{tabular}[c]{@{}c@{}}Aesthetic\\ Quality$\uparrow$\end{tabular}} & \multicolumn{1}{c}{\begin{tabular}[c]{@{}c@{}}Imaging\\ Quality$\uparrow$\end{tabular}} \\
\midrule
Moore-AnimateAnyone~\cite{AnimateAnyone} & 85.56 & 86.04 & 90.38 & 92.00 & 95.05 & 96.78 & 48.89 & 69.66    \\
HumanVid~\cite{HumanVid}            & 86.42 & 85.67 & 89.83 & 92.15 & 97.30 & 98.19 & 52.99 & 64.73    \\
MimicMotion~\cite{MimicMotion}         & 87.27 & 88.26 & 90.05 & 91.56 & 96.95 & 98.29 & 54.26 & 61.58    \\
Animate-X~\cite{Animate-X}           & 93.58 & 94.53 & 92.47 & 93.42 & 96.76 & 97.97 & 59.86 & 66.74    \\
UniAnimate-DiT~\cite{UniAnimate-DiT}      & 91.35 & 91.75 & 92.81 & 93.00 & 97.52 & 98.45 & 58.18 & 70.85    \\
VACE~\cite{vace}  & 75.92 & 78.29 & 89.51 & 92.15 & 97.00 & 98.01 & 53.01 & 60.74    \\
Wan-Animate~\cite{Wan-Animate}         & 88.71 & 89.23 & 90.65 & 92.67 & 97.41 & 98.40 & 56.57 & 62.79    \\
\rowcolor[rgb]{0.929,0.902,0.973}
\modelName~(Ours) & 96.17 & 96.92 & 91.61 & 93.38 & 97.10 & 98.37 & 61.57 & 71.74   \\
\bottomrule
\end{tabular}}
    \vfill 
    \caption{X-Dance dataset.}
    \label{tab:X_Dance}
  \end{subtable}
  \vspace{-8mm}
\end{table}

\begin{figure}[t]
    \centering
    \includegraphics[width=0.95\textwidth]{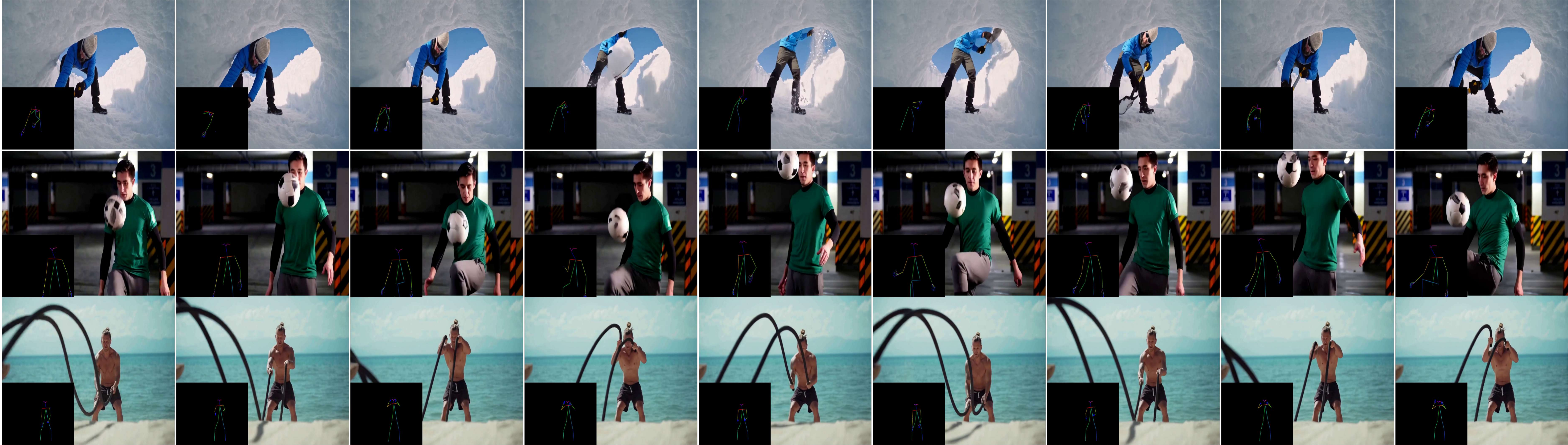}
    \caption{Visualization on RealisDance-Val about complex Human-Object Interactions. Even when driven solely by human pose, our model successfully synthesizes the interacting objects with \textbf{physically plausible motion and deformation}.}
    \label{fig:R-2}
    \vspace{-3mm}
\end{figure}

\begin{figure}[t]
    \centering
    \makebox[\textwidth][c]{%
        \begin{minipage}[t]{0.65\textwidth}
            \vspace{0pt}
            \centering
            \includegraphics[width=\linewidth]{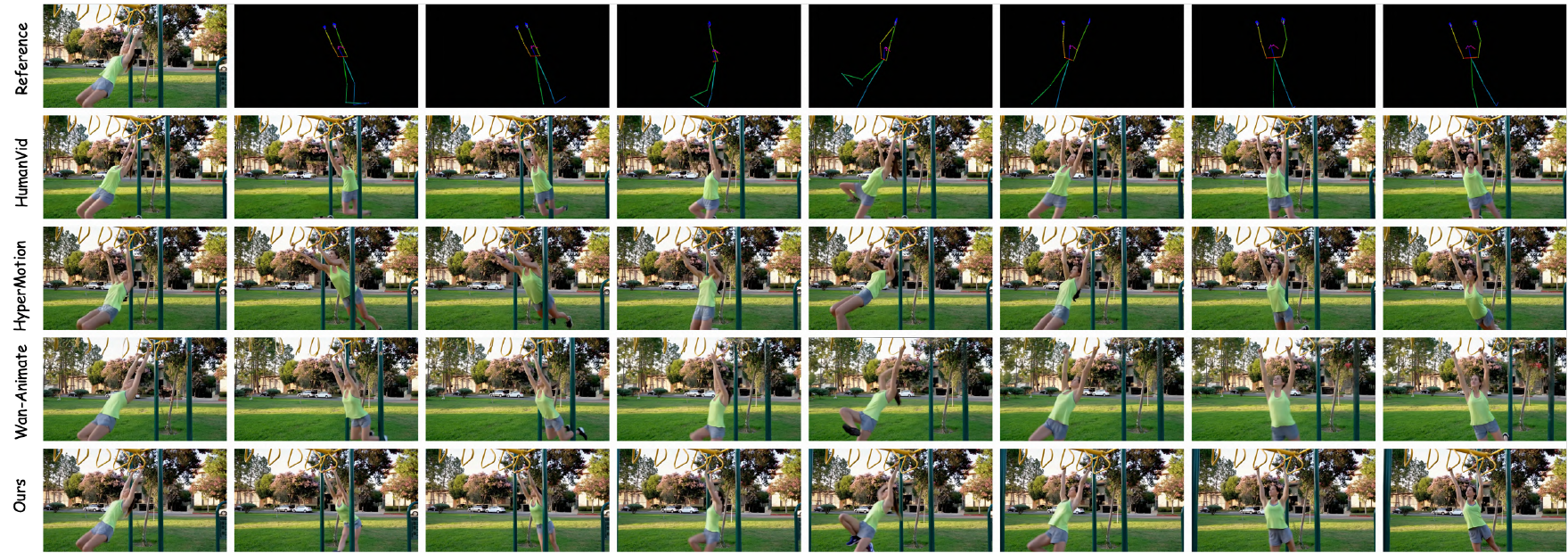}
            \caption{Comparison on RealisDance-Val, showing that our model achieves both precise control and \textbf{reasonable interaction with objects}, causing them to produce reasonable movements.}
            \label{fig:R-1}
            \vspace{1mm}
            \includegraphics[width=\linewidth]{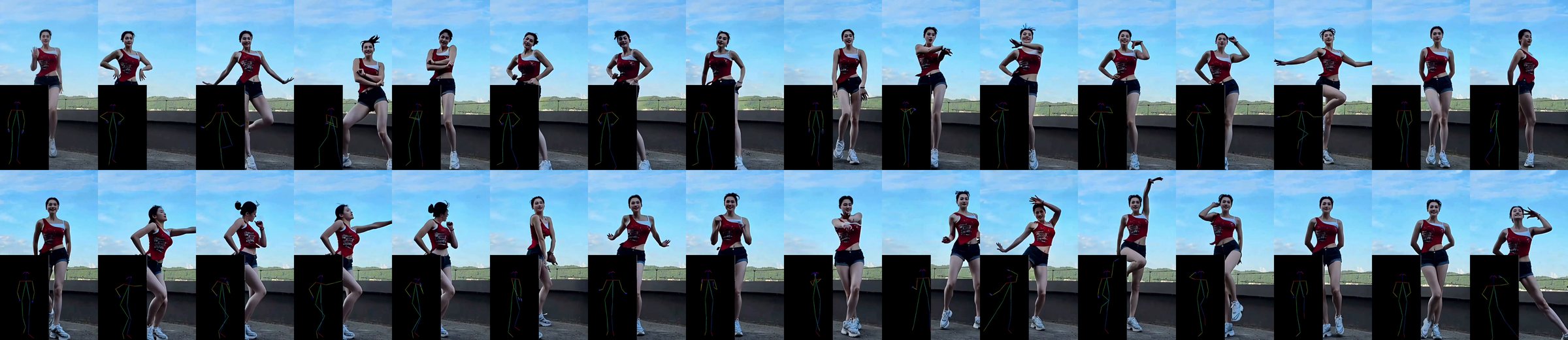}
            \caption{Training-free long-video generation result.}
            \label{fig:longvideo}
        \end{minipage}%
        \hspace{0.015\textwidth}%
        \begin{minipage}[t]{0.3\textwidth}
            \vspace{0pt}
            \centering
            \includegraphics[width=\linewidth]{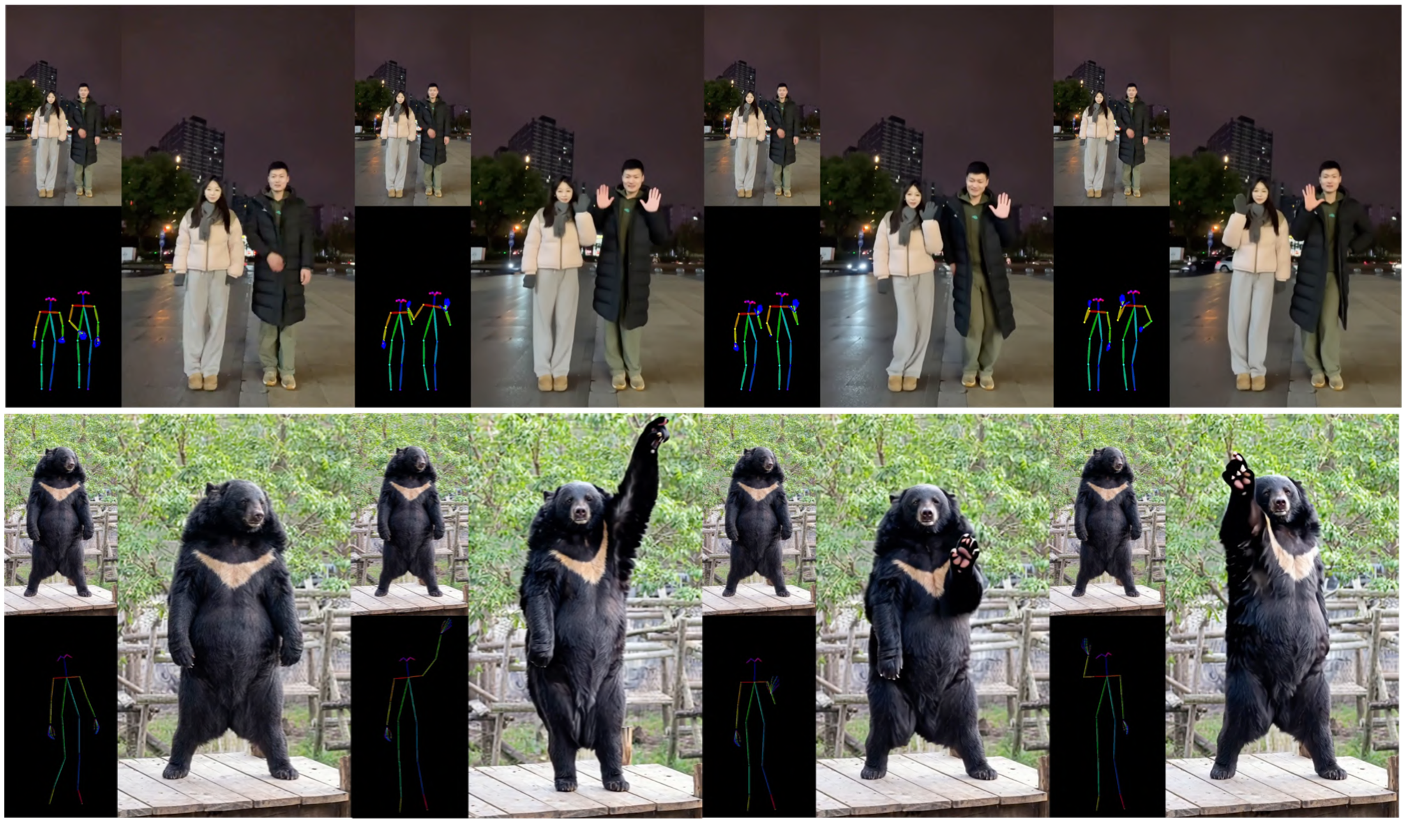}
            \caption{\textbf{Zero-shot generalization} to multi-person and animal scene.}
            \label{fig:R-3}
            \vspace{1mm}
            \includegraphics[width=\linewidth]{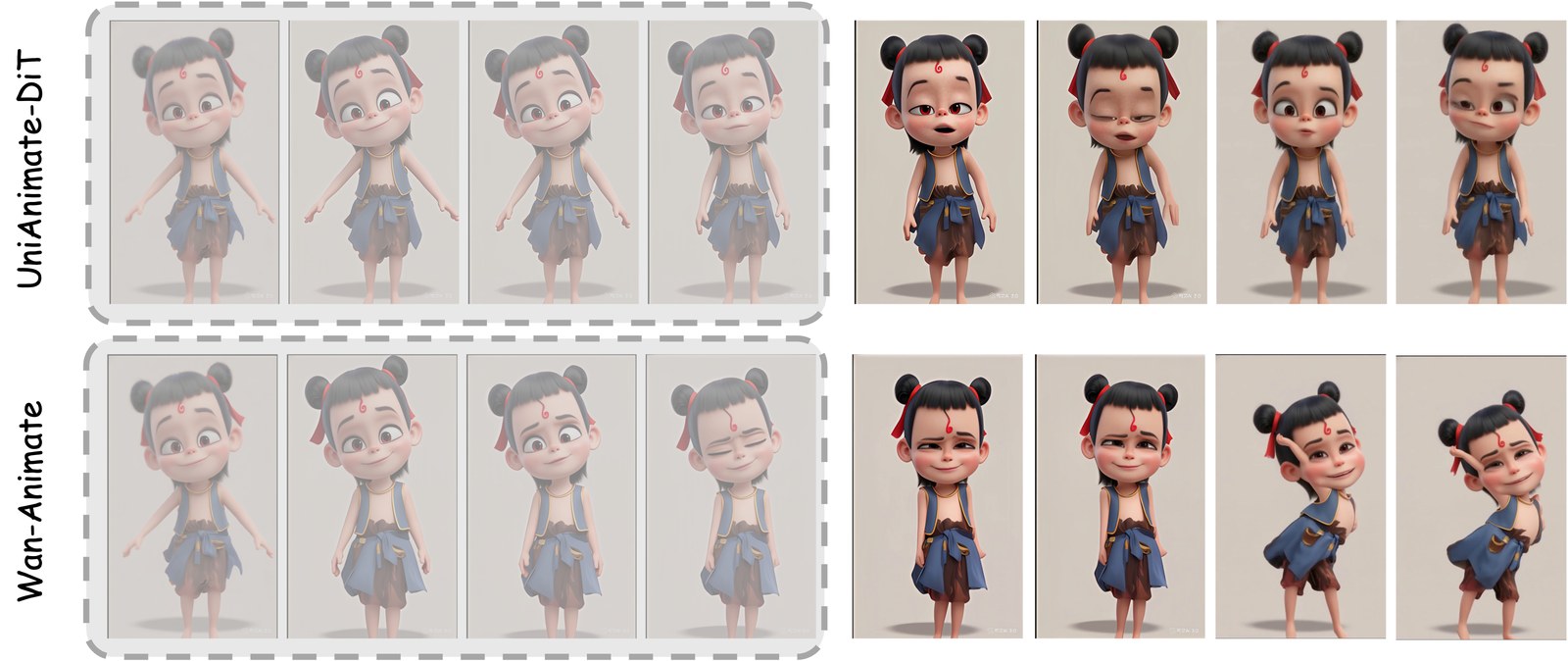}
            \caption{First-frame preservation with \textcolor{gray}{first-last-frame generation} post-processing.}
            \label{fig:flf2v}
        \end{minipage}%
    }
    \vspace{-4mm}
\end{figure}

\textbf{Implementation Details.}
We initialize from a pre-trained Wan-2.1 I2V 14B video model~\cite{Wan21} and conduct all experiments on 8 NVIDIA H800s.
Each sample contains 81 frames, with center resolution 640 for training and 768 for inference.
Our three stages run for 12,000, 2,000, and 500 steps, totaling only 14,500 steps.
As Table~\ref{tab:train_resource} shows, this is \textit{significantly fewer training steps} than other DiT-based methods, while still delivering strong motion control, quality, and smoothness.
Fig.~\ref{fig:ablation_ckpt} further shows that the first stage quickly learns control before focusing on details.

\noindent\textbf{Training Dataset.}
We collect a proprietary human motion dataset with 7,338 five-second clips, totaling 10.2 hours.
It mainly contains human dance sequences, with a few slow-motion documentary-style shots, and excludes extreme or complex movements.
As shown in Table~\ref{tab:train_resource}, \textit{the dataset scale is significantly smaller} than comparable works, demonstrating that the I2V paradigm leverages image priors to reduce training overhead over R2V.

\noindent\textbf{Evaluation Metrics.}
For the TikTok~\cite{jafarian2021learning_tiktok} dataset, we follow the settings in HumanVid~\cite{HumanVid}. The metrics consist of image quality (SSIM~\cite{wang2004image_ssim}, LPIPS~\cite{zhang2018unreasonable_lpips}, PSNR~\cite{hore2010image_psnr}, and FID~\cite{heusel2017gans_fid}) and video fidelity (FVD~\cite{unterthiner2018towards_fvd}).
For the RealisDance-Val~\cite{RealisDance-DiT} dataset, we utilize Vbench-I2V~\cite{huang2023vbench,huang2024vbench++} to facilitate fine-grained and objective evaluation.
For high-performance models, low-level metrics are insufficient proxies for generative quality, as they over-penalize minor shifts, miss semantic coherence, and are inapplicable without ground truth.
Although VBench provides a supplement, it lacks motion-driving accuracy measures; thus, qualitative visual effectiveness better reflects real-world performance, pending stronger metrics.

\subsection{Comparison with the State-of-the-Art methods}

\noindent\textbf{Quantitative results.}
As shown in Table~\ref{tab:tiktok_realisedance}, we evaluate TikTok with low-level metrics and RealisDance-Val/X-Dance with multi-dimensional Vbench-I2V metrics.
SteadyDancer achieves highly competitive results, especially on representative FID, FVD, and VBench metrics, including our real-world-oriented, non-homogeneous X-Dance benchmark.

\noindent\textbf{Qualitative comparisons.}
To evaluate spatio-temporal misalignments beyond same-source benchmarks, we propose \textit{\textbf{\dataName}}, a \textit{different-source} benchmark with diverse references (male/female/cartoon and upper-/full-body shots) and challenging driving videos with blur, occlusion, \textit{spatial-structural inconsistencies}, and \textit{temporal start-gaps}.
As shown in Fig.~\ref{fig:compare_I2V}, competing methods often fail to preserve identity or follow driving motion under these challenges, whereas our model maintains first-frame identity with precise, coherent motion control.
We also evaluate \textit{\textbf{RealisDance-Val}}, which includes daily/dance motions and \textit{complex Human-Object Interactions} that test pose following and interaction potential.
As shown in Fig.~\ref{fig:R-2} and Fig.~\ref{fig:R-1}, our model synthesizes interacting objects with \textit{physically plausible motion and deformation} while preserving appearance, unlike competing methods that often produce static artifacts or shape collapse.

\noindent\textbf{Generalization to Multi-person and Animal Scenarios.}
We further explore our model's zero-shot generalization in multi-person (with a multi-person pose estimator) and animal scenarios. As shown in Fig.~\ref{fig:R-3}, our model seamlessly handles multi-person pose inputs, maintaining the same performance of identity preservation and motion fidelity as in single-person. Remarkably, our framework achieves this performance without any multi-person/animal training data or specialized inference-time tuning.

\noindent\textbf{Training-free Long-Video Generation.}
Although our default training and inference setting uses 81 frames, SteadyDancer can be extended to minute-long video generation in a training-free manner using overlapping context-window inference, with 81-frame sliding windows, FreeNoise initialization, and linear overlap fusion.
As shown in Fig.~\ref{fig:longvideo}, the generated long video maintains strong identity preservation and coherent motion over an extended duration.

\noindent\textbf{First-frame preservation without post-processing.}
R2V methods can also achieve first-frame preservation by first-last-frame post-processing, which uses the reference frame and generated first frame to synthesize \textcolor{gray}{the shaded segment} in Fig.~\ref{fig:flf2v}, but this adds computation and \emph{still fails to preserve the reference appearance} during the animation generation part.
In contrast, SteadyDancer achieves first-frame preservation intrinsically through I2V generation, making it not only a native capability but also an effective means of preserving identity.

\begin{table}[t] 
    \centering
    \captionsetup{skip=1pt}
    \begin{minipage}[t]{0.42\textwidth}
        \vspace{0pt}
        \centering
        \includegraphics[width=0.84\linewidth]{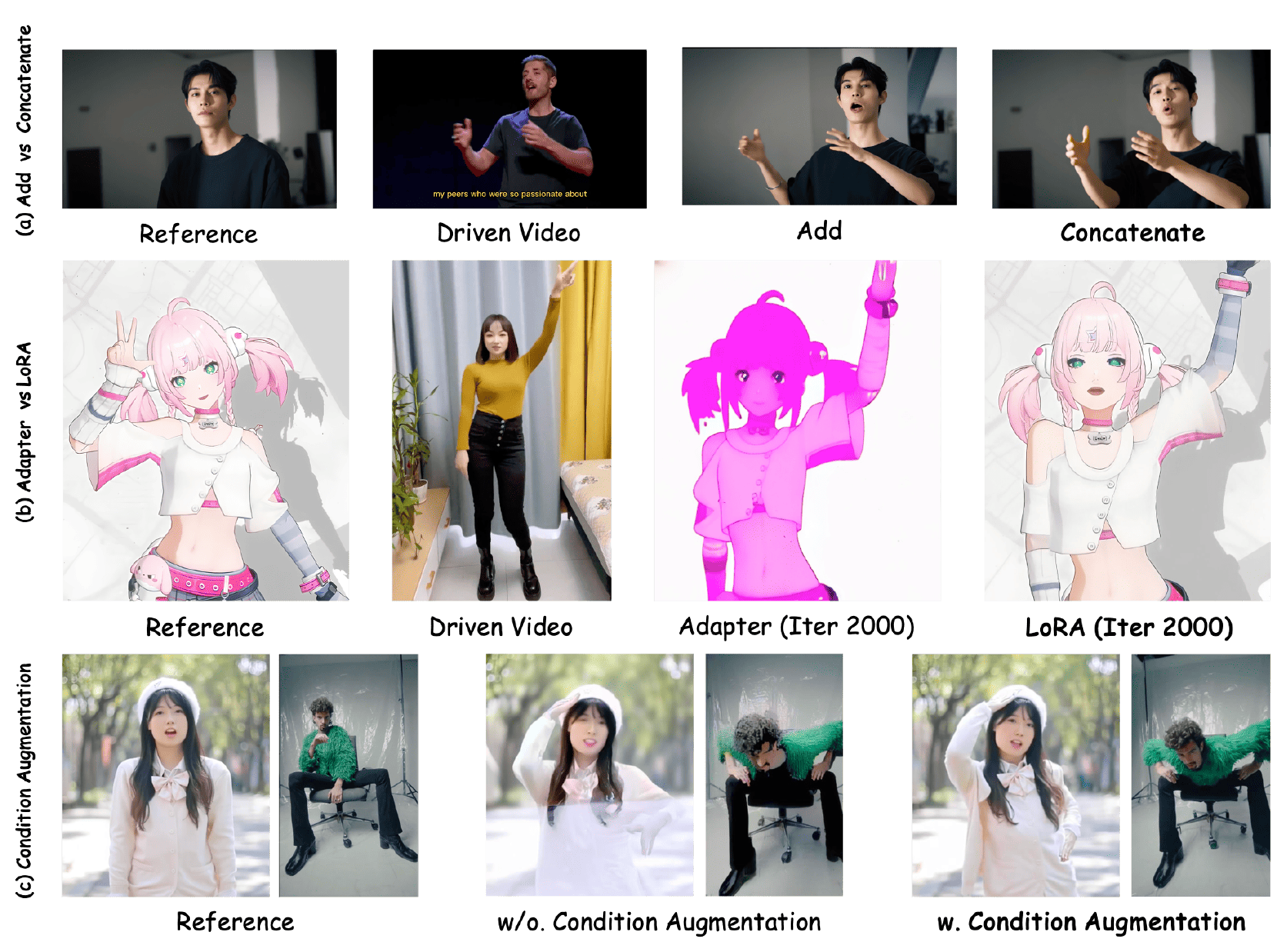}
        \vspace{0.5mm}
        \captionof{figure}{Ablation on Condition-Reconciliation Mechanism.}
        \label{fig:abblation_ref}
    \end{minipage}
    \hfill
    \begin{minipage}[t]{0.55\textwidth}
        \centering
        \vspace{5pt}
        \caption{Ablation on Synergistic Pose Modulation Modules and the training stage.}
        \label{tab:ablation}
        \renewcommand{\arraystretch}{1.5}
        \resizebox{\linewidth}{!}{\begin{tabular}{l|ccccc|cccc}
\toprule
\multicolumn{1}{c|}{\multirow{2}{*}{Settings}} & \multicolumn{5}{c|}{RealisDance-Val}    & \multicolumn{4}{c}{X-Dance}   \\ \cline{2-10} 
\multicolumn{1}{c|}{}             & IS $\uparrow$    & IB $\uparrow$    & SC $\uparrow$    & TF $\uparrow$    & FVD $\downarrow$    & IS $\uparrow$    & IB $\uparrow$    & SC $\uparrow$    & TF $\uparrow$    \\
\midrule
Stage 1      & 94.49 & 95.09 & 93.24 & 97.00 & 330.47 & 93.22 & 93.70 & 87.74 & 96.46 \\
Stage 1 w/o SSAE   & 94.48 & 95.09 & 93.17 & 96.34 & 349.98 & 92.69 & 94.65 & 87.41 & 96.35 \\
Stage 1 w/o FFAU               & 94.36 & 95.08 & 93.21 & 96.06 & 339.55 & 92.44 & 93.91 & 87.74 & 96.41 \\
Stage 1 w/o TMCM   & 94.37 & 95.01 & 93.24 & 96.19 & 333.52 & 92.25 & 93.41 & 87.82 & 96.43 \\ \midrule
Stage 2 w C-D Distillation             & 94.66 & 95.28 & 93.31 & 97.56 & 329.55 & 93.44 & 94.11 & 87.74 & 96.51 \\
Stage 2 w Normal Distillation      & 79.13 & 81.10 & 88.98 & 97.12 & 723.91 & -     & -     & -     & -     \\ \midrule
Stage 3      & 94.68 & 95.38 & 93.48 & 97.99 & 326.49 & 96.17 & 96.92 & 91.61 & 97.10 \\
\bottomrule
\end{tabular}}
    \end{minipage}
    \vspace{-7mm}
\end{table}

\subsection{Ablation Study}

\noindent \textbf{Condition-Reconciliation Mechanism.}
To achieve precise control with minimal data while preserving I2V priors, condition integration is crucial.
As shown in Fig.~\ref{fig:abblation_ref}, channel concatenation better preserves identity than element-wise addition (Row 1) or adapter-based injection (Row 2), and removing Condition Augmentation (Row 3) degrades appearance fidelity.

\begin{figure}[t]
    \centering
    \captionsetup{skip=1pt}
    \hspace*{\fill}%
    \begin{minipage}[c]{0.40\textwidth}
        \centering
        \includegraphics[width=0.84\linewidth]{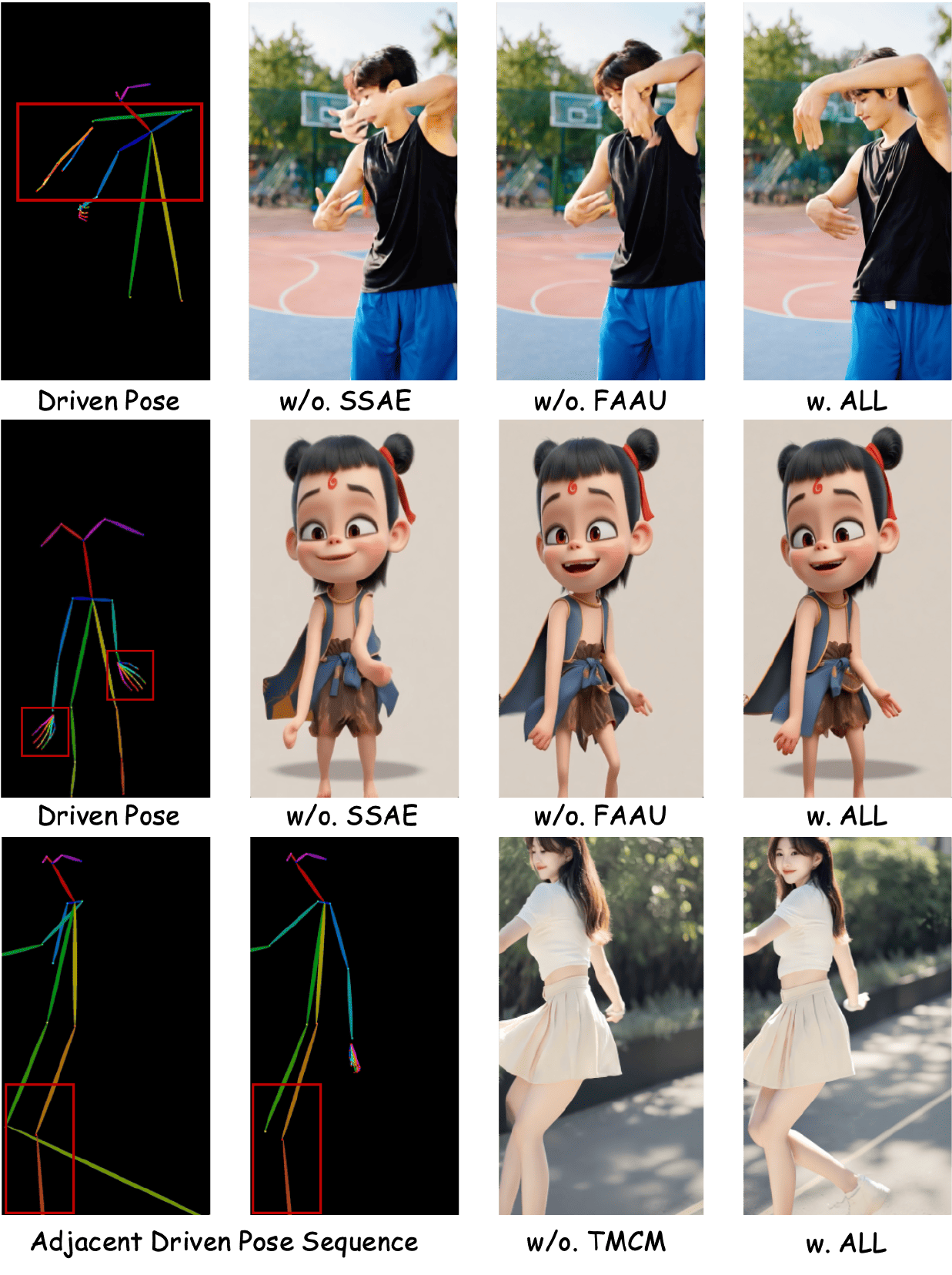}
        \vspace{0.5mm}
        \caption{Ablation on Synergistic Pose Modulation Modules.}
        \label{fig:ablation_motion_alignment}
    \end{minipage}
    \hspace*{\fill}%
    \begin{minipage}[c]{0.48\textwidth}
        \centering
        \begin{minipage}{\linewidth}
            \centering
            \includegraphics[width=\linewidth]{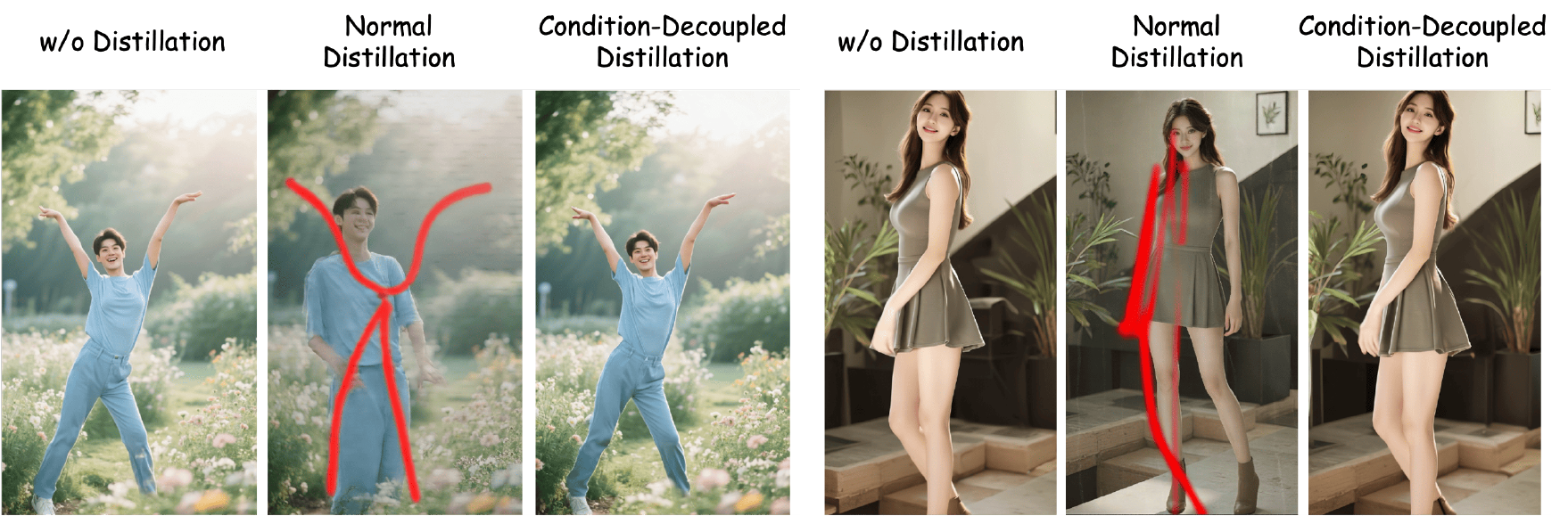}
            \vspace{-2mm}
            \caption{Ablation on Condition-Decoupled Distillation. Red lines are incorrectly artifacts in Normal Distillation.}
            \label{fig:ablation_cdd}
        \end{minipage}
        
        \vspace{-2mm}
        
        \begin{minipage}{\linewidth}
            \centering
            \includegraphics[width=\linewidth]{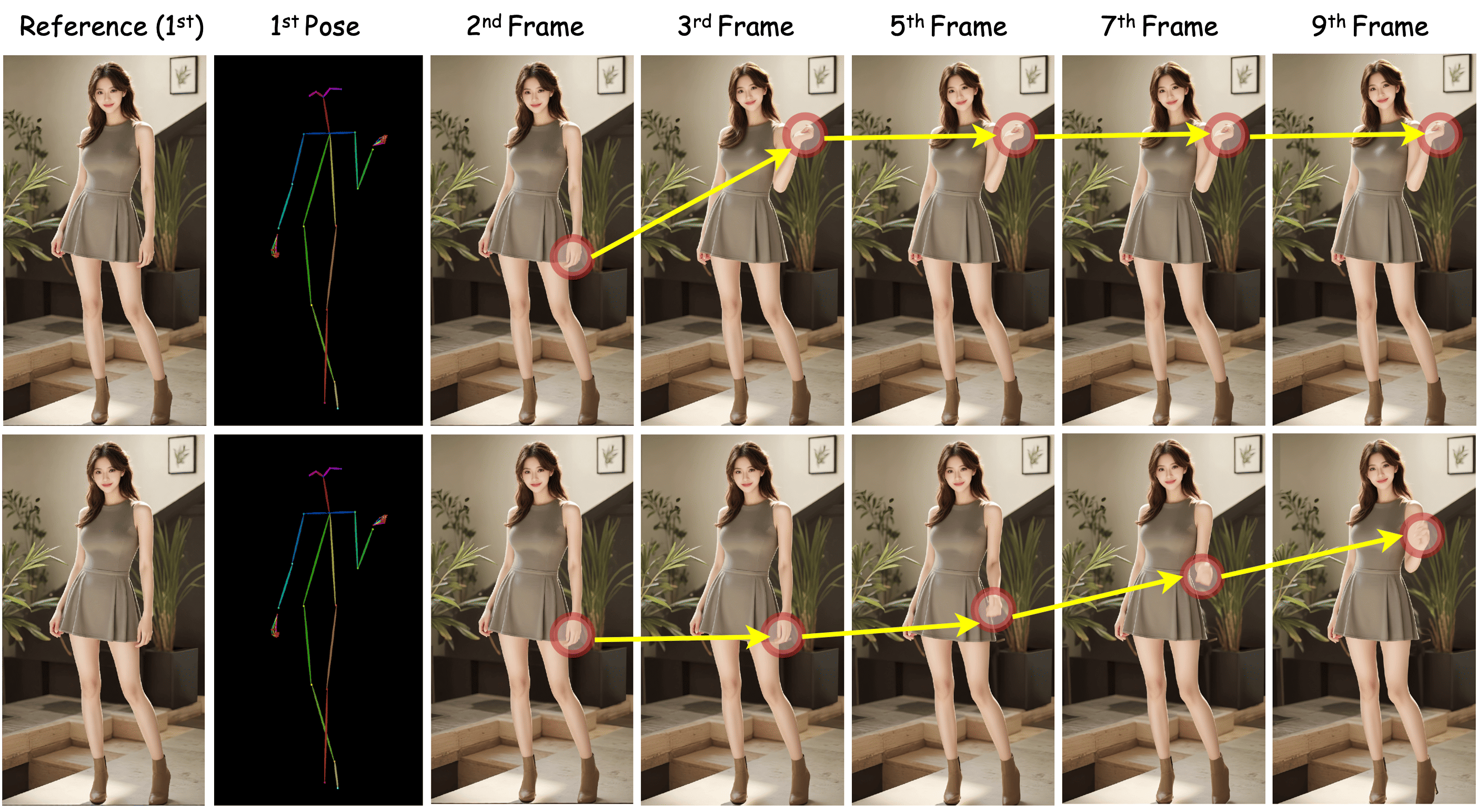}
            \vspace{-2mm}
            \caption{Ablation study on Motion Discontinuity Mitigation.}
            \label{fig:ablation_mdas}
        \end{minipage}
    \end{minipage}%
    \hspace*{\fill}
    \vspace{-8mm}
\end{figure}

\noindent \textbf{Synergistic Pose Modulation Modules.}
Robust Motion-to-Image Alignment remains challenging in I2V due to pose inaccuracies and spatio-temporal misalignments, and Fig.~\ref{fig:ablation_motion_alignment} validates the distinct roles of our modules.
For spatial issues, SSAE handles pose errors (Row 1) and FAAU addresses large structural disparities (Row 2) with multi-scale and reference-adaptive representations.
For temporal issues, TMCM models missing or contradictory poses (Row 3) for smooth guidance, with VBench results in Table~\ref{tab:ablation} further confirming all three modules.

\noindent\textbf{Condition-Decoupled Distillation.}
To recover generation quality after motion-control learning, we introduce Condition-Decoupled Distillation in Sec.~\ref{sec:method-training-cdd} and compare it with the first-stage model and normal MSE distillation.
As shown in Figure~\ref{fig:ablation_cdd}, our design improves video fidelity while maintaining stability, whereas conventional distillation collapses.
This failure stems from conflicts between unconditional distillation gradients and pose-conditional optimization, which gradually desensitize the model to conditional inputs.
By decoupling these objectives, our method mitigates inter-branch interference, as further supported by Table~\ref{tab:ablation}.

\noindent\textbf{Motion Discontinuity Mitigation.}
To address start-gap misalignment between the reference image and first pose frame, we propose Pose Simulation in Sec.~\ref{sec:method-training-mdas}.
As shown in Figure~\ref{fig:ablation_mdas}, this discrepancy causes an abrupt baseline artifact between the second and third frames.
In contrast, our Pose Imitation fine-tuning synthesizes the hand-raising motion as a smooth, plausible transition.
This validates its ability to mitigate motion discontinuities, also reflected by the Temporal Flickering gains in Table~\ref{tab:ablation}.

\section{Conclusion}

We present SteadyDancer, a framework for harmonized, coherent human animation that leverages first-frame preservation. It resolves the core I2V challenge of harmonizing fidelity with motion control and ensuring coherence via our novel Condition-Reconciliation Mechanism and Synergistic Pose Modulation Modules.
Our Staged Decoupled-Objective Training pipeline efficiently optimizes for motion, quality, and continuity with minimal resources. Quantitative and qualitative results, especially on our X-Dance benchmark, show SteadyDancer significantly outperforms competitors. We believe these innovations provide a solid, efficient method for future robust human animation.

\newpage
{
    \small
    \bibliographystyle{ieeenat_fullname}
    \bibliography{neurips_2026}
}


\newpage
\appendix
\noindent\textbf{\Large{Appendix}}

\vspace{3mm}

\section{X-Dance}
\begin{figure}[!t]
    \centering
    \includegraphics[width=\textwidth]{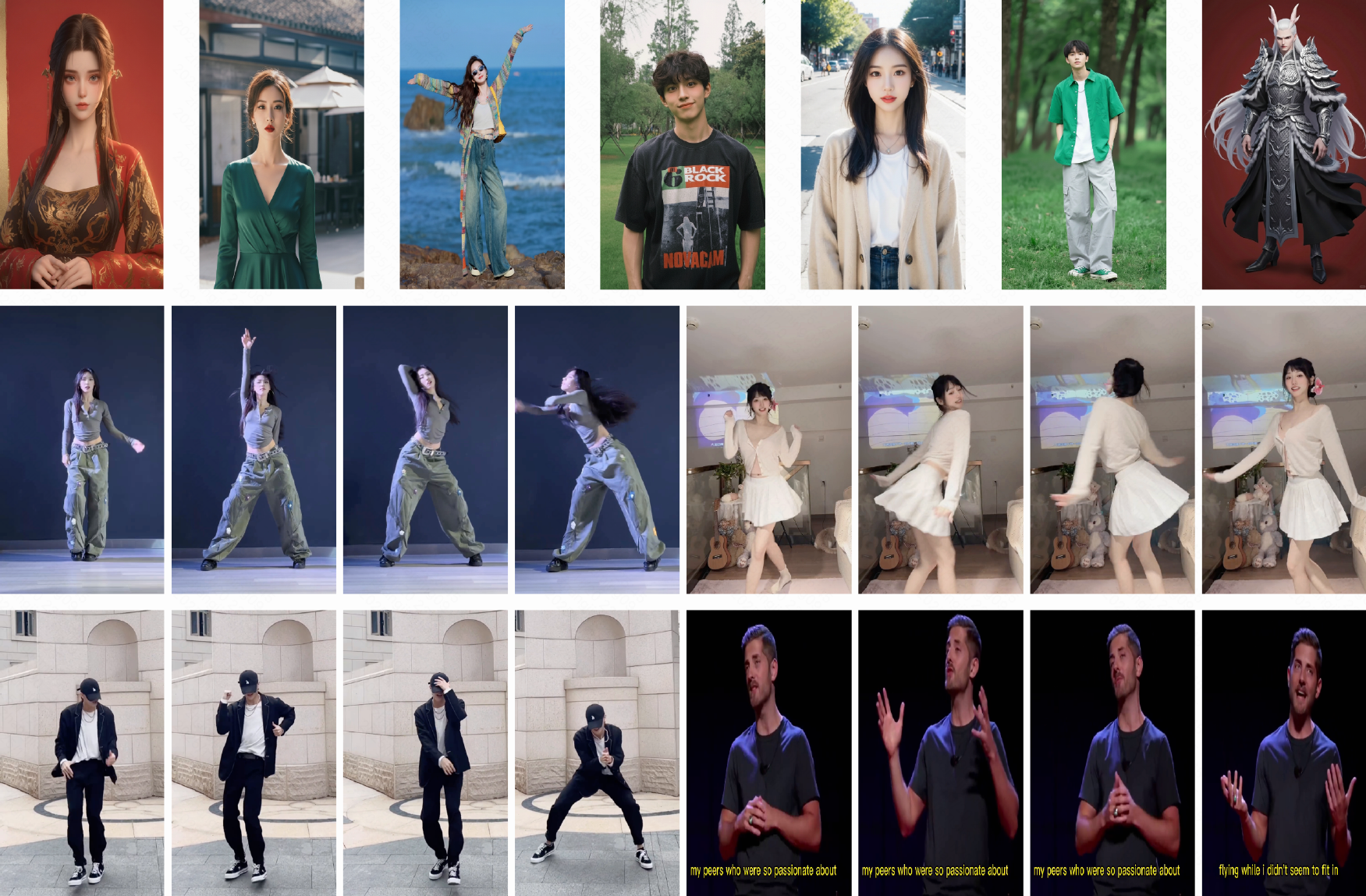}
    \vspace{-1mm}
    \caption{Examples from the X-Dance benchmark. The second and third rows display driving video sequences, comprising both intricate, high-dynamic dance movements and low-amplitude simple activities. The first row presents reference images, which were specifically curated relative to these driving videos to evaluate the model with real-world misalignment challenges.}
    \label{fig:X-Dance}
    \vspace{-2mm}
\end{figure}

Standard benchmarks, such as TikTok and RealisDance, source both the reference image and pose sequence from the same video. This idealized setup fails to reflect the spatio-temporal misalignment challenges prevalent in real-world applications.
As shown in Fig.~\ref{fig:X-Dance}, to more robustly evaluate the model's generalization capabilities in such scenarios, we curated and introduced a new evaluation dataset, \dataName.
We first collected 12 distinct driving videos, comprising 8 sequences of intricate, high-dynamic dance movements and 4 sequences of low-amplitude daily activities. These sequences are replete with non-ideal real-world factors, such as motion blur, severe occlusion, and drastic pose changes.
Tailored to these motions, we specifically curated a diverse set of reference images to simulate real-world misalignments. This specially designed collection contains: (1) anime characters to introduce stylistic domain gaps; (2) half-body shots to represent compositional inconsistencies; (3) cross-gender or anime characters to simulate significant skeletal structural discrepancies; and (4) subjects in distinct postures to maximize the initial action gap.
By systematically pairing these reference images with the 12 driving videos, we simulate two critical real-world challenges: (1) Spatial pose-structure inconsistency (e.g., an anime character driving a real-world pose); and (2) Temporal discontinuity, specifically the significant gap between the reference pose and the initial driving pose.

\begin{figure}[t]
    \centering
    \includegraphics[width=\textwidth]{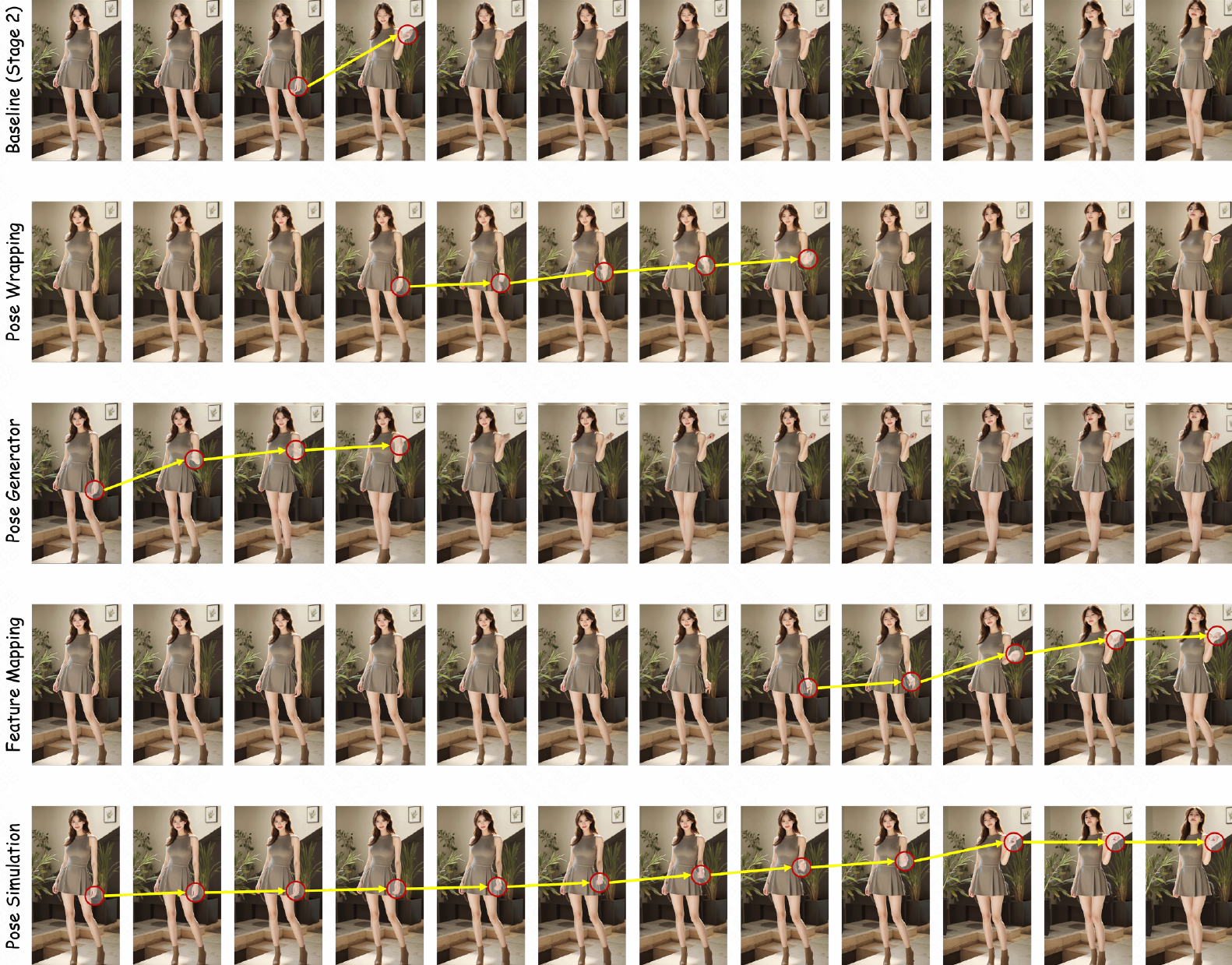}
    \vspace{-1mm}
    \caption{Performance comparison of four Motion Discontinuity Mitigation methods, showing that the Pose Simulation approach generates smooth and natural transitions. Notably, this method achieves this without introducing additional modules or extra inference latency.}
    \label{fig:stage3_1_4_compare}
    \vspace{-2mm}
\end{figure}

\begin{figure}[t]
    \centering
    \includegraphics[width=\textwidth]{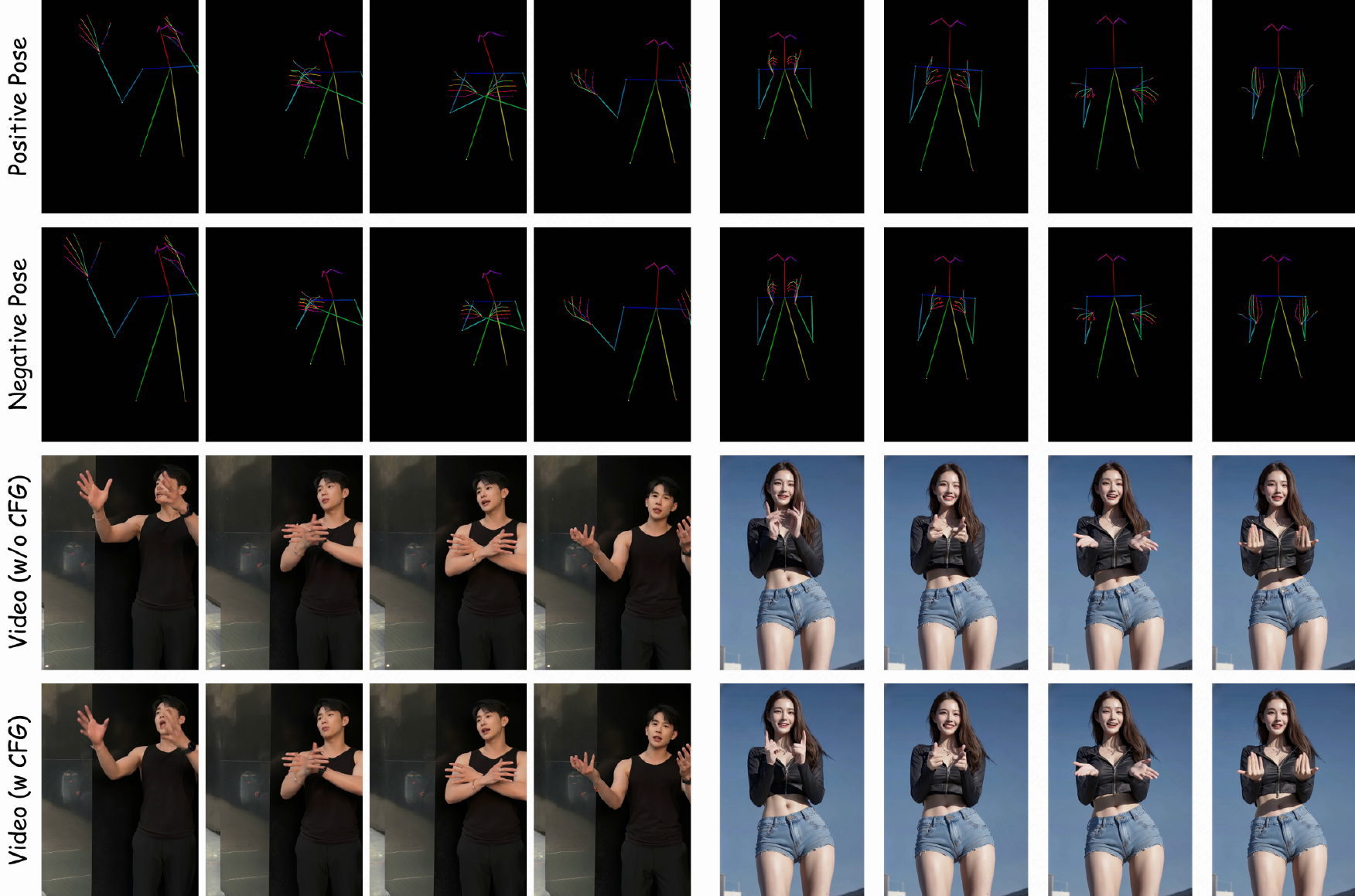}
    \vspace{-1mm}
    \caption{Model performance using Decoupled-Condition Classifier-Free Guidance (DC-CFG). From top to bottom, the rows display the original pose (positive condition), the perturbed pose (negative condition), generation results without DC-CFG, and generation results with DC-CFG, showing our DC-CFG improves pose control and effectively suppresses generation artifacts.}
    \label{fig:cfg_compare}
    \vspace{-2mm}
\end{figure}

\section{Model Details}

\subsection{Motion Discontinuity Mitigation.}
As discussed in the main text, to address the abrupt transition between the reference frame and the initial pose frame, we propose \textbf{Pose Simulation} to explicitly replicate this discontinuity within the training data.
Specifically, given a smooth training sequence $\{p_{0}, p_{1}, \dots, p_{T}\}$, we first construct synthetic pairs $(p_{0}, p_{T^{*}})$, where the target timestamp $T^{*}$ is randomly sampled from $\{2, 3, 4\}$ to ensure diversity.
We then interpolate between this pair to generate intermediate poses, selecting the first interpolated frame $\tilde{p}_{1}$. With a probability of 0.5, we replace the original $p_{1}$ with this synthetic $\tilde{p}_1$, yielding a pseudo-training sample $\{p_{0}, \tilde{p}_1, p_{2}, \dots, p_{T}\}$.
This strategy effectively mimics realistic discontinuities while preserving the integrity of the motion control signal. By directly utilizing these synthetic trajectories as training samples, the model is exposed to realistic jump patterns without requiring any architectural modifications.
Notably, fine-tuning on these synthetic samples for just a few hundred steps resolves over 80\% of extreme jump scenarios, all while maintaining no additional module parameters and zero additional inference latency.

Beyond our proposed strategy, we explored several alternative approaches to address the motion discontinuity:
\begin{itemize}
    \item \textbf{Pose Warping}: We inserted an explicit pose-interpolation submodule designed to generate intermediate poses bridging the reference pose and the first driving frame, effectively implementing pose warping.
    \item \textbf{Pose Generator}: We devised a lightweight pose-sequence generator trained to synthesize intermediate frames conditioned on the start and end poses.
    \item \textbf{Feature Mapping}: We designed a feature-mapping module that enhances each pose latent by aggregating features from its adjacent temporal neighbors.
\end{itemize}
As shown in Fig.~\ref{fig:stage3_1_4_compare}, a comparison of the four methods reveals that Pose Simulation yields the best results, producing smooth and natural transitions superior to the alternatives.
Furthermore, all three approaches necessitate the introduction of additional network modules, which proved difficult to optimize effectively given our limited training data. In contrast, Pose Simulation mitigates this issue from a data-centric perspective. It is perfectly suited to our limited data regime and imposes zero additional parameter overhead.

\subsection{Implementation Details.}

\noindent\textbf{Training Details.}
As detailed in the main text, our Staged Decoupled-Objective Training Pipeline employs distinct training configurations to optimize different objectives across separate stages, all leveraging the same dataset.
\begin{itemize}
    \item Stage 1: Action Supervision. We utilize LoRA-based training with a learning rate of 1e-4 for 12,000 steps. Model selection in this phase prioritizes motion adherence and basic image quality.
    \item Stage 2: Condition-Decoupled Distillation. We switch to full-parameter fine-tuning with a reduced learning rate of 1e-6 for 2,000 steps. The selection of the optimal checkpoint is governed by visual preference.
    \item Stage 3: Motion Discontinuity Mitigation. We revert to LoRA-based training with a learning rate of 1e-4 for a brief 500 steps, focusing on leveraging the augmented training data to mitigate motion discontinuity artifacts.
\end{itemize}

\noindent\textbf{Training Dataset.}
We curated a proprietary human motion dataset consisting of 7,338 five-second video clips, totaling 10.2 hours. Notably, this dataset scale is significantly smaller than that of comparable works, highlighting the data efficiency of our design.
To ensure quality, these clips were rigorously filtered from the Internet based on aesthetic scores, motion smoothness, subject prominence, and action types. The final collection is composed of two parts: $\sim$2,000 clips sourced from high-quality footage (e.g., documentaries, YouTube) to introduce diversity in aspect ratios and motion dynamics; and $\sim$5,000 vertical videos collected primarily from social media, focusing predominantly on dance sequences. We purposefully excluded extreme or complex actions to maintain training stability.

\begin{figure}[t]
    \centering
    \includegraphics[width=0.9\textwidth]{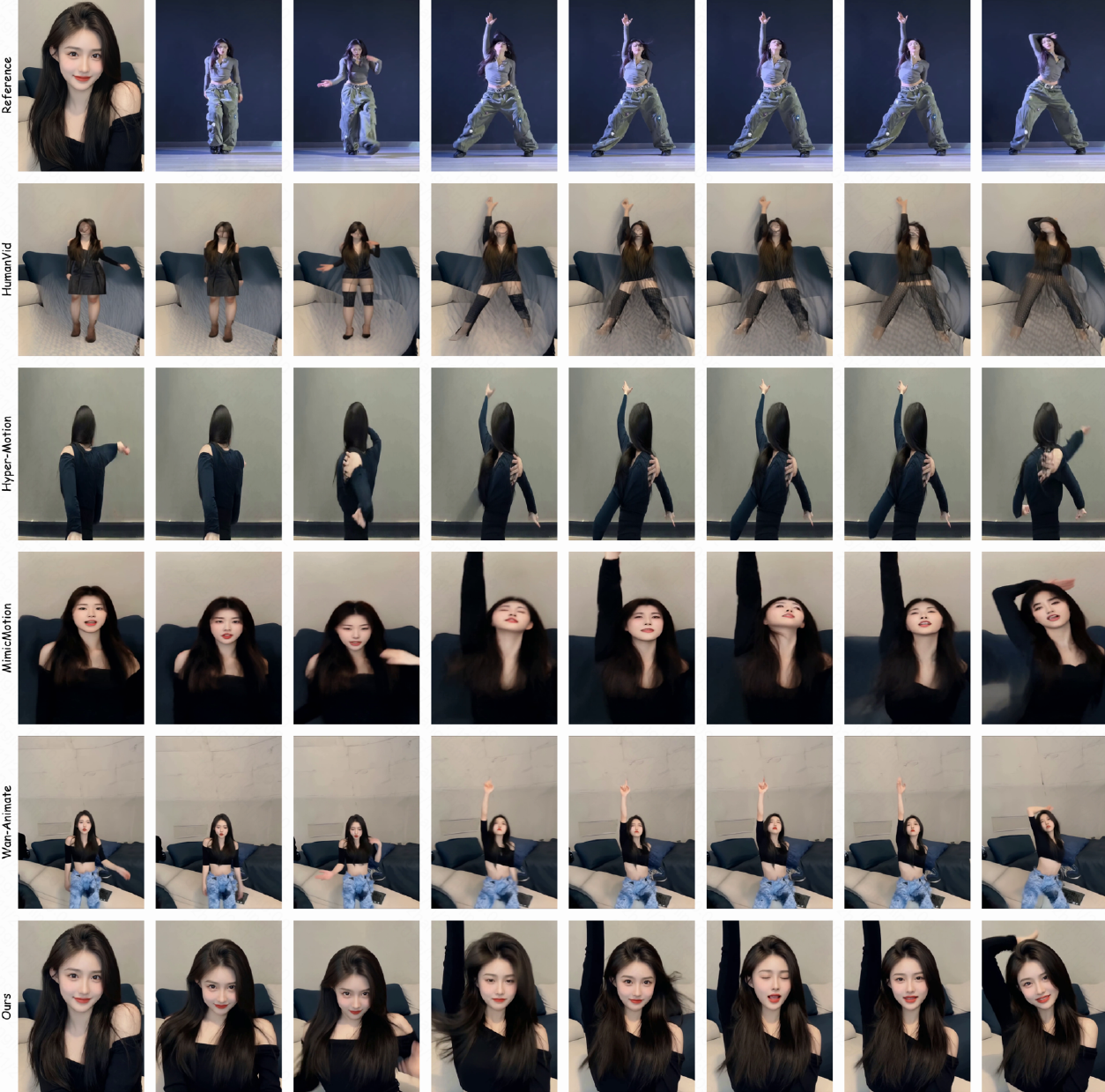}
    \vspace{-1mm}
    \caption{More visualization comparison on X-Dance.}
    \label{fig:X-3}
    \vspace{-2mm}
\end{figure}

\begin{figure}[t]
    \centering
    \includegraphics[width=0.9\textwidth]{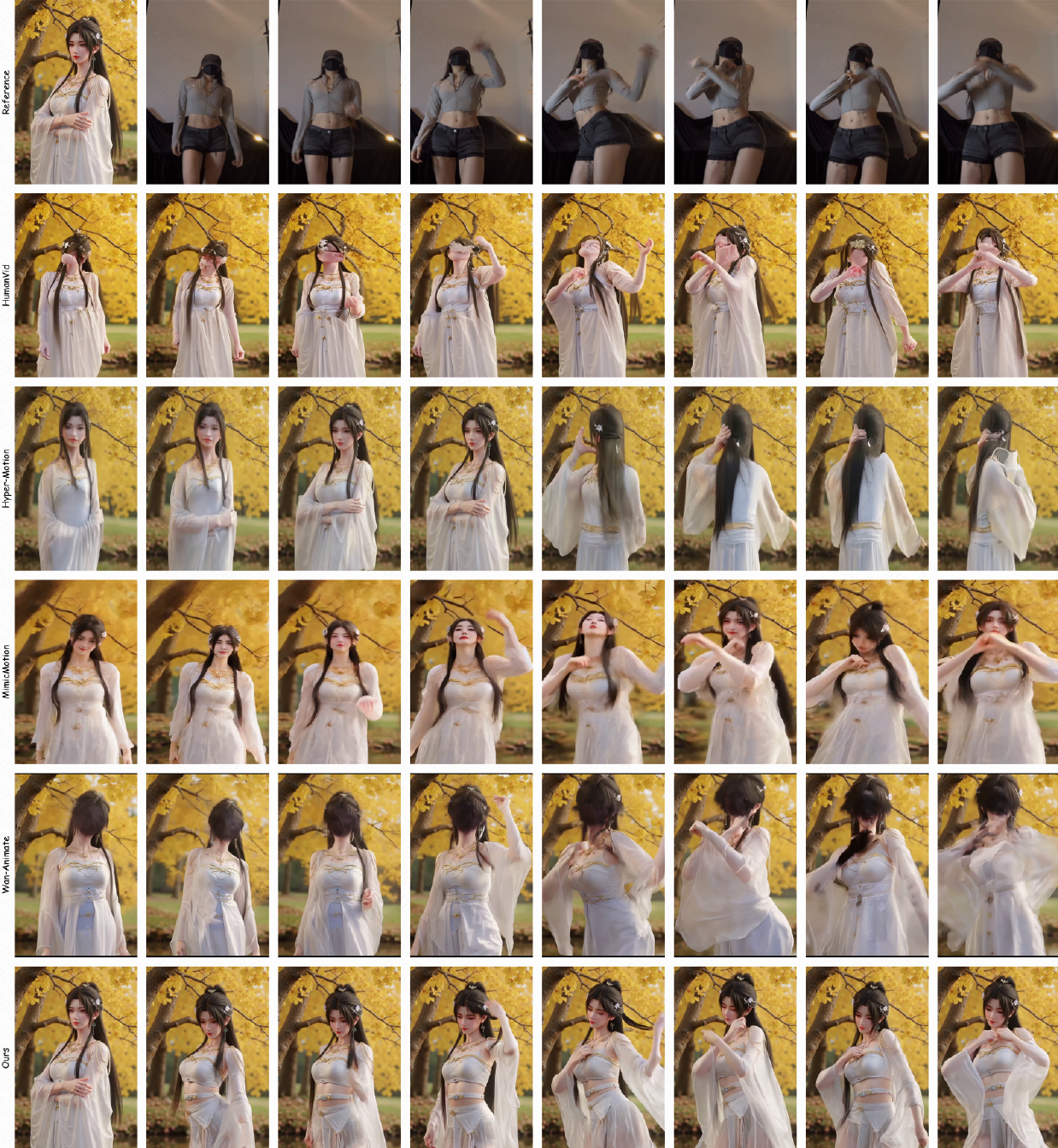}
    \vspace{-1mm}
    \caption{More visualization comparison on X-Dance.}
    \label{fig:X-2}
    \vspace{-2mm}
\end{figure}

\begin{figure}[t]
    \centering
    \includegraphics[width=0.9\textwidth]{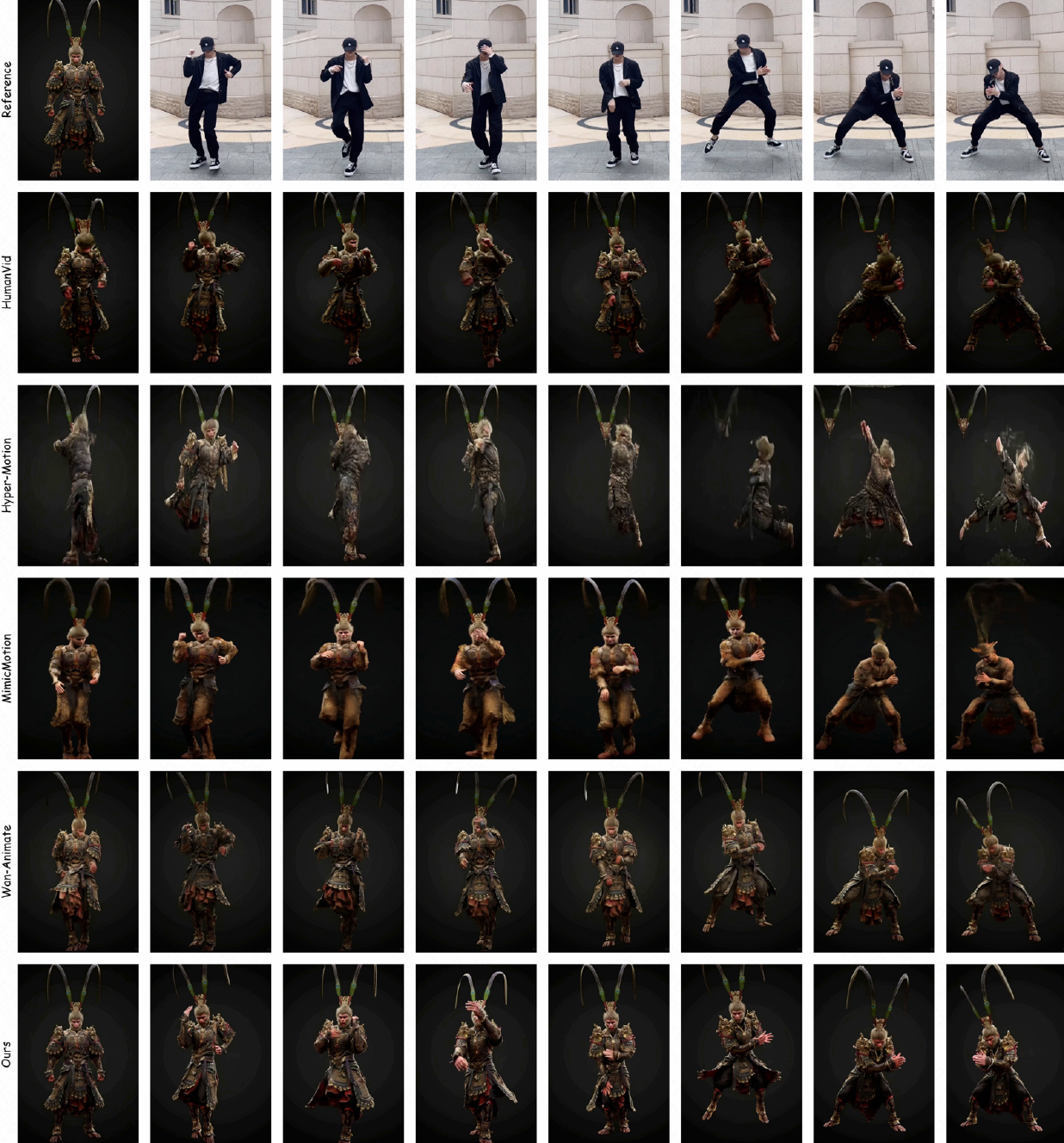}
    \vspace{-1mm}
    \caption{More visualization comparison on X-Dance.}
    \label{fig:X-1}
    \vspace{-2mm}
\end{figure}

\subsection{Decoupled-Condition Classifier-Free Guidance.}
Most of the existing video generation models typically employ Classifier-Free Guidance (CFG) to synthesize high-quality samples that strictly adhere to the provided conditional guidance. 
Specifically, during the sampling process, the model leverages its capability to predict both conditional and unconditional noise.
At each denoising timestep $t$, we perform two forward passes through the DiT using the current noisy latent $x_t$, one conditioned on $y$ ($\epsilon_{\theta}(x_t, t, y)$), and another conditioned on a null embedding $\emptyset$ ($\epsilon_{\theta}(x_t, t, \emptyset)$).
The standard CFG noise prediction is formulated as:
\begin{equation}
\begin{aligned}
    \hat{\epsilon}_{\theta}(x_t, t, y, w) &= \epsilon_{\theta}(x_t, t, \emptyset) \\
    + & w(\epsilon_{\theta}(x_t, t, y) - \epsilon_{\theta}(x_t, t, \emptyset))    
\end{aligned}
\end{equation}
where the difference between the two predictions represents a vector in noise space that steers the generation towards the condition $y$. The scalar $w$ denotes the guidance scale, determining the strength of the shift from the unconditional distribution toward the conditional one.
In practice, it is common to replace the null condition $\emptyset$ with a specific negative prompt ($y_{neg}^{txt}$) to provide a more explicit negative constraint, thereby improving generation quality.

Within our framework, the pose signal $y^{pose}$ serves as a critical condition. Inspired by textual negative prompting, we propose an innovative Decoupled-Condition Classifier-Free Guidance (DC-CFG) to further enhance pose controllability.
Specifically, we apply \textit{scale and shift perturbations} to the extracted pose signal to construct a negative pose condition $y_{neg}^{pose}$.
This explicitly simulates the misalignment issues we aim to avoid. Based on this, we obtain a prediction guided by the negative pose: $\epsilon_{\theta}(x_t, t, y_{neg}^{pose})$. To effectively integrate these multiple conditions, we decouple the guidance and adjustment processes as follows:
\begin{equation}
\begin{aligned}
    \Delta \epsilon_{\theta}(x_t, t, y, y_{neg}^{pose}) &= \epsilon_{\theta}(x_t, t, y) - \epsilon_{\theta}(x_t, t, y_{neg}^{pose}) \\
    \Delta \epsilon_{\theta}(x_t, t, y, y_{neg}^{txt}) &= \epsilon_{\theta}(x_t, t, y) - \epsilon_{\theta}(x_t, t, y_{neg}^{txt}) \\
    \hat{\epsilon}_{\theta}(x_t, t, y, w^{pose}, w^{txt}) &= \epsilon_{\theta}(x_t, t, y_{neg}^{txt}) \\
    + w^{pose} \cdot \Delta \epsilon_{\theta}(x_t, t, y, & y_{neg}^{pose}) + w^{txt} \cdot \Delta \epsilon_{\theta}(x_t, t, y, y_{neg}^{txt}) \\
\end{aligned}
\end{equation}
where $w^{pose}$ and $w^{txt}$ denote the guidance scales for the pose and text prompt, respectively. By leveraging a pose-conditioned CFG, we compel the generator to not only strictly adhere to the target pose but also actively diverge from imprecise or ambiguous neighboring poses. This mechanism significantly sharpens motion precision, effectively suppressing common artifacts such as limb blurring, motion ghosting, or regression to the mean pose, ultimately achieving higher-fidelity control of the driving motion.

Furthermore, we carefully designed a temporal scheduling strategy for pose-guided CFG, capitalizing on the inherent coarse-to-fine generation characteristic of diffusion models.
The core philosophy is to modulate the guidance strength dynamically.
In the early denoising stages, the model primarily constructs low-frequency components, such as global contours and spatial layout. Imposing strong guidance at this phase ensures that the character's global pose structure is precisely anchored.
Conversely, as the process advances to the later stages, the model shifts its focus to rendering high-frequency details, such as texture and lighting. Here, it is crucial to attenuate or remove the guidance. This grants the model sufficient degrees of freedom to synthesize photorealistic details faithful to the source appearance, avoiding visual artifacts like structural rigidity or unnatural deformations caused by over-guidance.
This temporal mechanism effectively decouples structural pose control from appearance detail generation. By maximizing realism and naturalness while ensuring motion precision, it serves as a pivotal trade-off strategy for achieving high-fidelity pose-driven video generation.
In practice, we apply pose-conditioned CFG exclusively within the normalized timestep interval of $[0.1, 0.4]$. The guidance scales are configured as $w^{pose}=1.0$ for the pose condition and $w^{txt}=5.0$ for the text prompt.
Fig.~\ref{fig:cfg_compare} presents the positive and perturbed negative pose conditions along with generation results, illustrating the impact of our DC-CFG. As shown, our approach improves pose control and effectively suppresses generation artifacts.

\subsection{Efficiency Comparison}

To illustrate the comparison in terms of efficiency, we also provide a comparative analysis among Wan-based methods in the Table~\ref{tab:gpu}. Moreover, we support multi-GPU parallel inference and provide a near-lossless FP16 version integrated with LoRA acceleration from LightX2V to facilitate practical deployment. We will also release quantized models to further reduce overhead.

\begin{table}[!t]
\caption{Efficiency comparison among Wan-based Model.}
\label{tab:gpu}
  \centering
  \small
  \setlength{\tabcolsep}{1pt}
  \renewcommand\arraystretch{1}
   \resizebox{0.8\linewidth}{!}{%
    \begin{tabular}{c|c|cc|cc}
\toprule
\multirow{2}{*}{Model}         & \multirow{2}{*}{Params (B)} & \multicolumn{2}{c|}{Single GPU} & \multicolumn{2}{c}{Two-GPU parallel} \\ \cline{3-6} 
                               &                             & 81 frame (min)   & ~Memory (GB)   & 81 frame (min)      & ~Memory (GB)     \\
\midrule
UniAnimate-DiT                 & 16.40                       & 20.5             & 24           & -                   & -              \\
VACE                           & 17.33                       & OOM              & OOM          & 11.7                & 79             \\
Wan-Animate                    & 17.27                       & 11.9             & 62           & 6.7                 & 56             \\
\rowcolor[rgb]{0.929,0.902,0.973}
Ours                           & 16.39                       & 17.2             & 43           & 10.2                & 41             \\
\rowcolor[rgb]{0.929,0.902,0.973}
Ours$_\text{FP16, LoRA acceleration}^\text{offload, block swap}$ & 16.39                           & 3.7                & 9.2            & -                   & -             \\
\bottomrule
\end{tabular}

  }
\end{table}

\section{More Qualitative Results}
We present qualitative comparisons between our method and state-of-the-art approaches, including HumanVid~\cite{HumanVid}, Hyper-Motion~\cite{HyperMotion}, MinicMotion~\cite{MimicMotion}, and Wan-Animate~\cite{Wan-Animate}, on our challenging X-Dance dataset in Fig.~\ref{fig:X-3}, Fig.~\ref{fig:X-2}, and Fig.~\ref{fig:X-1}. Notably, our method not only achieves superior generation quality with first-frame preservation but also requires significantly fewer training resources.

\section{Limitation and Future Work}
Despite the promising results achieved by SteadyDancer in harmonized and coherent animation, several limitations remain to be addressed.
1) Domain Gap in Stylized Images. While our model delivers visually pleasing and coherent results for anime reference frames, its performance remains slightly inferior to the exceptional fidelity achieved on real-world images, occasionally exhibiting minor degradation in stylistic consistency. This limitation stems from the under-representation of anime samples in our current training corpus. In future work, we aim to augment the training data with specialized anime datasets to bridge this domain gap and enhance the model's generalization across different artistic styles.
2) Extreme Motion Discontinuities. Our proposed strategy effectively mitigates motion discontinuities in the vast majority of scenarios. However, in cases of extreme pose discrepancies between the reference frame and the initial driving pose, as the model prioritizes precise motion control, it may generate transitions that appear accelerated or slightly unnatural. We believe that developing more sophisticated temporal modeling architectures and scaling up the training data will be instrumental in further resolving this start-gap challenge.
3) Advanced Motion Representation. The current architecture relies heavily on the accuracy of the input pose sequence; for instance, consecutive pose estimation errors can lead to irreversible artifacts in the generated video. We argue that an ideal animation system should balance precise controllability with high error tolerance. Therefore, designing a more refined, diverse, and semantically rich motion representation remains a promising direction for future research.

\section{Broader Impact}
SteadyDancer can benefit creative applications such as film production, advertising, game development, virtual content creation, and digital human animation by reducing the resources required for high-fidelity motion-driven video generation.
However, as with other human image animation and video generation technologies, it may also be misused to animate a person's image without consent or to create misleading synthetic media.
We therefore encourage responsible use of the released code and model resources, including obtaining appropriate consent for identity-driven generation, avoiding deceptive or harmful applications, and following applicable data, copyright, and platform usage policies.
The released code and model resources will be provided under a usage license that prohibits unauthorized identity animation, deceptive synthetic media generation, and other harmful applications.

\section{Existing Assets}
We use and cite existing models, datasets, benchmarks, and evaluation tools, including Wan-2.1, TikTok, RealisDance-Val, VBench/VBench++, and baseline methods.
These assets are used for research and evaluation purposes following their respective licenses and terms of use.
Our proprietary training data are not released and are used under applicable data usage restrictions.



\end{document}